\documentclass[10pt,journal,compsoc]{IEEEtran}
\usepackage{graphicx}
\usepackage{booktabs}
\usepackage{amsmath}
\usepackage{subcaption}
\usepackage{wrapfig}
\usepackage{multirow}
\usepackage{amssymb}
\usepackage{booktabs}
\usepackage{bbm}
\usepackage{xcolor}
\usepackage{adjustbox}
\definecolor{LightPurple}{rgb}{0.88,0.88,1}
\definecolor{LightCyan}{rgb}{0.88,1,1}
\definecolor{Gray}{gray}{0.85}
\usepackage{colortbl}
\usepackage{arydshln}
\usepackage{threeparttable}
\usepackage{textgreek} 
\newcommand{\minisection}[1]{\vspace{0.005in} \noindent {\bf #1}}
\usepackage{ragged2e}
\usepackage{bbding}

\ifCLASSOPTIONcompsoc
  \usepackage[nocompress]{cite}
\else
  \usepackage{cite}
\fi
\ifCLASSINFOpdf
\else
\fi

\newcommand\MYhyperrefoptions{bookmarks=true,bookmarksnumbered=true,
pdfpagemode={UseOutlines},plainpages=false,pdfpagelabels=true,
colorlinks=true,linkcolor={black},citecolor={black},urlcolor={black},
pdftitle={Bare Demo of IEEEtran.cls for Computer Society Journals},%<!CHANGE!
pdfsubject={Typesetting},%<!CHANGE!
pdfauthor={Michael D. Shell},%<!CHANGE!
pdfkeywords={Computer Society, IEEEtran, journal, LaTeX, paper,
             template}}%<^!CHANGE!
\ifCLASSINFOpdf
\usepackage[\MYhyperrefoptions,pdftex]{hyperref}
\else
\usepackage[\MYhyperrefoptions,breaklinks=true,dvips]{hyperref}
\usepackage{breakurl}
\fi

\begin{document}

\title{Exploiting the Semantic Knowledge of Pre-trained Text-Encoders for Continual Learning}

\author{Lu~Yu, 
        Zhe~Tao,
        Dipam~Goswami, 
        Hantao~Yao,~\IEEEmembership{Member,~IEEE,}
        Bartłomiej~Twardowski,
        Joost~Van~de~Weijer,        and~Changsheng~Xu,~\IEEEmembership{Fellow,~IEEE}% <-this % stops a space
\IEEEcompsocitemizethanks{
\IEEEcompsocthanksitem Lu Yu and Zhe Tao are with School of Computer Science and Engineering, Tianjin University of Technology, 300384, China (e-mail: luyu@email.tjut.edu.cn, tz@stud.tjut.edu.cn).
\IEEEcompsocthanksitem D. Goswami, B. Twardowski, J. van de Weijer are with the Computer Vision Center, Universitat Autonoma de Barcelona, Barcelona 08193, Spain. (e-mail: dgoswami@cvc.uab.es, btwardowski@cvc.uab.cat, joost@cvc.uab.es).
\IEEEcompsocthanksitem Hantao Yao is with School of Information Science and Technology, University of Science and Technology of China, Hefei, 230026, China (e-mail: yaohantao@ustc.edu.cn).
\IEEEcompsocthanksitem Changsheng Xu is the corresponding author, with State Key Laboratory of Multimodal Artificial Intelligence Systems, Institute of Automation, Chinese Academy of Sciences, Beijing, 100190, China, and also with the School of Artificial Intelligence, University of the Chinese Academy of Sciences, Beijing, 100049, China (e-mail: csxu@nlpr.ia.ac.cn).
}% <-this % stops a space
\thanks{Manuscript received August, 2024; revised May 30, 2025.}}

\markboth{Journal of \LaTeX\ Class Files,~Vol.~X, No.~X, May~2025}%
{Shell \MakeLowercase{\textit{et al.}}: Bare Advanced Demo of IEEEtran.cls for IEEE Computer Society Journals}

\IEEEtitleabstractindextext{%
\begin{abstract}\justifying
  Deep neural networks (DNNs) excel on fixed datasets but struggle with incremental and shifting data in real-world scenarios. Continual learning addresses this challenge by allowing models to learn from new data while retaining previously learned knowledge.  
  Existing methods mainly rely on visual features, often neglecting the rich semantic information encoded in text. 
  The semantic knowledge available in the label information of the images, offers important semantic information that can be related with previously acquired knowledge of semantic classes.    
  Consequently, effectively leveraging this information throughout continual learning is expected to be beneficial.
  To address this, we propose integrating semantic guidance within and across tasks by capturing semantic similarity using text embeddings. We start from a pre-trained CLIP model, employ the \emph{Semantically-guided Representation Learning (SG-RL)} module for a soft-assignment towards all current task classes, and use the \emph{Semantically-guided Knowledge Distillation (SG-KD)} module for enhanced knowledge transfer. Experimental results demonstrate the superiority of our method on general and fine-grained datasets. \emph{Our code can be found in \href{https://github.com/aprilsveryown/semantically-guided-continual-learning}{https://github.com/aprilsveryown/semantically-guided-continual-learning}.}
\end{abstract}

}

\maketitle

\IEEEdisplaynontitleabstractindextext
\IEEEpeerreviewmaketitle

\ifCLASSOPTIONcompsoc
\IEEEraisesectionheading{\section{Introduction}\label{sec:introduction}}
\else
\section{Introduction}
\label{sec:introduction}
\fi

\IEEEPARstart{C}ontinual learning (CL) aims to enable models to learn sequentially from a stream of tasks without catastrophically forgetting previously acquired knowledge. A large variety of algorithms were proposed to mitigate forgetting, employing approaches such as rehearsal~\cite{castro2018end,hou2019learning,rebuffi2017icarl,zhao2020maintaining}, regularization~\cite{kirkpatrick2017overcoming, li2017learning,zhao2023rethinking}, and architectural modifications~\cite{douillard2020podnet, douillard2022dytox, serra2018overcoming, yan2021dynamically,madaan2023heterogeneous}. The emergence of Vision Transformers (ViTs)~\cite{dosovitskiy2020image} has substantially enhanced the representation capabilities of pre-trained vision encoders for downstream tasks. A number of recent studies, including L2P~\cite{wang2022learning} and DualPrompt~\cite{wang2022dualprompt}, have leveraged the potential of pre-trained vision encoders in the context of continual learning, employing a prompt-based learning strategy. However, these works have exposed a notable limitation: performance degradation occurs when the pre-trained dataset exhibits a substantial semantic gap compared to the downstream data~\cite{tang2023prompt, zhang2023slca}. 

Large scale pre-trained vision-language models (VLMs), for instance ViT-BERT~\cite{zhou2020unified}, CLIP~\cite{radford2021learning} and UNITER~\cite{chen2020uniter}, have emerged as powerful tools that combine computer vision and natural language processing, enabling machines to comprehend visual and textual information~\cite{yutext,khattak2023maple}. These models can serve as strong foundation models for continual learning, as they were trained on massive datasets, allowing them to learn intricate patterns and relationships between images and language. As mentioned, the continual learning community has 
focused on applying foundational vision backbones~\cite{wang2022learning, wang2022dualprompt,smith2023coda,roy2024convolutional,zhou2024expandable}. However, how to best exploit the rich semantic knowledge embedded within the language encoders of VLMs is an active and evolving area of investigation. Early efforts like Continual-CLIP~\cite{thengane2022clip} demonstrated the potential by utilizing both vision and language encoders of CLIP, albeit by keeping the entire model frozen, which can limit adaptability to new tasks. 
Language guidance from pre-trained language encoders has been introduced into prompt-based continual learning methods, such as LGCL~\cite{khan2023introducing}. This approach aligns the task selection mechanism in the prompt pool with the overall language description of the task and aligning the visual features with the language description of specific classes. 

\begin{figure}[t]
    \centering
    \begin{subfigure}{\linewidth}
        \centering
        \begin{minipage}{\linewidth}
            \centering
            \resizebox{\linewidth}{!}{%
            \begin{tabular}{l|ccc}
            \toprule
            Method & Language & Vision & Learnable Backbone \\ \midrule
            DER \cite{yan2021dynamically} & x & scratch & yes \\
            DyTox++\cite{douillard2022dytox} & x & scratch & yes \\
            CODA-Prompt\cite{smith2023coda} & x & pre-trained & no \\
            EASE \cite{zhou2024expandable} & x & pre-trained & no \\
            ConvPrompt \cite{roy2024convolutional} & x & pre-trained & no \\
            RAPF \cite{huang2024class} & pre-trained & pre-trained & no\\
            CLAP4CLIP \cite{jhaclap4clip} & pre-trained & pre-trained & no\\
            Ours & pre-trained & pre-trained & yes \\ \bottomrule
            \end{tabular}}
            \caption{}
        \end{minipage}
    \end{subfigure}
    \begin{subfigure}{0.48\textwidth}
        \centering
        \begin{minipage}{\linewidth}
            \centering
            \includegraphics[width=\linewidth]{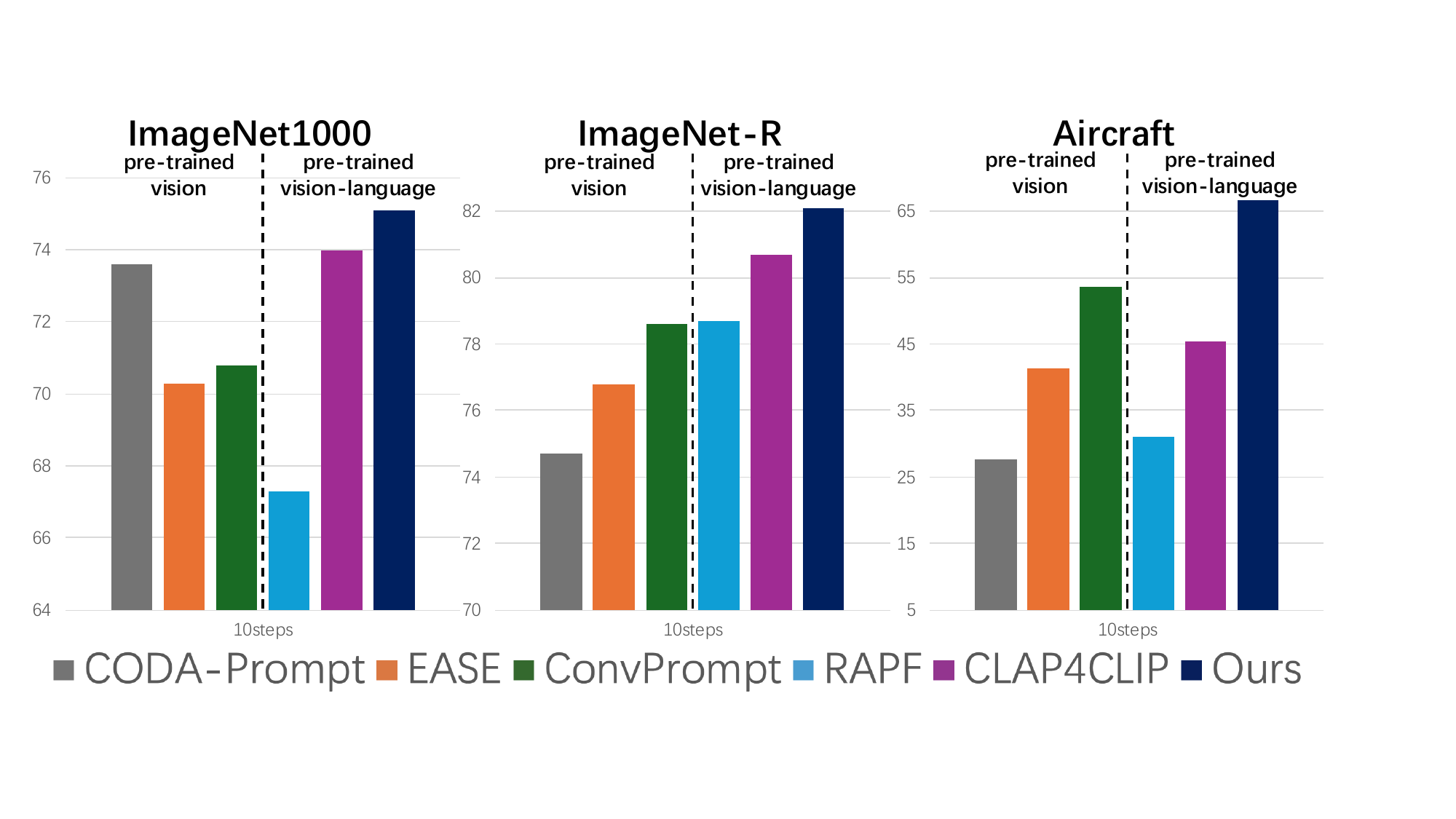} 
            \caption{}
        \end{minipage}
    \end{subfigure}%
    \caption{(a) Comparison of different models. (b) Performance comparison using different models across three datasets.}
   \label{fig:intro}
\end{figure}

Establishing similarity between objects is a fundamental aspect of human perception and cognition, as highlighted by previous research~\cite{shepp2013object}. Pre-trained text encoders provide access to the semantic similarity between classes.
In this paper, we propose two techniques to exploit the rich semantic information of pre-trained text encoders. Firstly, we exploit the \emph{intra-task semantic relationships} between class labels. Instead of only using the available ground truth label to train on the newly arriving data, we replace the label with a soft-assignment towards all current task classes based on intra-task semantic relations. This ensures that each sample contributes to aligning the vision backbone with respect to multiple semantic classes.
Secondly, we propose to exploit the \emph{inter-task semantic relationships} between old and new classes to improve stability during continual learning. For example, current class data for `motor bikes' is expected to assign part of its prediction to the related previous task class of `bikes' thereby preventing forgetting of this class. These inter-task relationships, derived from the language encoder, are used in a semantically-guided knowledge distillation technique. 

In Fig.~\ref{fig:intro}(a), we present an overview of several methods, each characterized by a distinct training strategy. The performance evaluations are shown in Fig.~\ref{fig:intro}(b).
We observe that the incorporation of pre-trained vision-language models 
leads to 
performance improvements
on ImageNet-R, which poses a large semantic gap from CLIP’s pre-training distribution. This suggests that the rich semantic priors from language supervision enhance the model's generalization to out-of-distribution categories. However, methods with frozen backbones, such as CLAP4CLIP~\cite{jhaclap4clip}, exhibit clear limitations when facing fine-grained and domain-shifted datasets like Aircraft, likely due to their inability to adapt feature representations to new tasks. In contrast, our approach effectively combines the benefits of vision-language pre-training with end-to-end adaptability, achieving superior performance across all datasets and outperforming both vision-only and frozen vision-language baselines under challenging class-incremental learning scenarios.

To summarize, the main contributions of the paper are:
\begin{itemize}
    \item 
    We demonstrate that effectively leveraging the semantic knowledge from pre-trained language encoders, in conjunction with vision encoders, significantly enhances performance in class-incremental learning, moving beyond the predominant focus on vision-only pre-training or less direct uses of language information.

    \item We introduce the \emph{SG-RL} module, which leverages intra-task semantic similarity between class labels to replace conventional one-hot label supervision. By aligning features with multiple semantically related class embeddings rather than treating classes as mutually independent, this module encourages the model to capture richer semantic structure within each task, leading to more discriminative and transferable representations.
    
    \item We extend the existing image distillation method based only on visual information with a term that exploits knowledge of the labels of the current data with respect to the classes of previous tasks (inter-task semantic similarity). This improvement, called \emph{SG-KD},  significantly increases the efficiency of distillation (and reduces catastrophic forgetting). 

    \item 
    Extensive experimental validation across a wide range of benchmarks—including the standard CIFAR100, 
    large-scale datasets like ImageNet-subset and ImageNet-1000, 
    eight diverse fine-grained datasets 
    and challenging few-shot learning scenarios 
    —confirms that our approach consistently delivers state-of-the-art or highly competitive performance.
\end{itemize}
 
\section{Related Work}

\subsection{Continual Learning}
Continual learning research has witnessed significant advancements in recent years, 
early approaches~\cite{rebuffi2017icarl, kirkpatrick2017overcoming,li2017learning,yu2020semantic,zhu2021prototype,yu2022self,liu2022long, tao2024class} can be broadly divided into three categories: rehearsal-based, regularization-based, and architecture-based. Rehearsal-based methods store previous task exemplars or generate fake exemplars using generative models to combat catastrophic forgetting. iCaRL~\cite{rebuffi2017icarl} combined exemplar rehearsal with knowledge distillation, UCIR~\cite{hou2019learning} addressed classification bias through normalization, and WA~\cite{zhao2020maintaining} ensured consistency in weight vector norms between old and new classes. Regularization-based methods impose constraints on network parameters, with strategies such as estimating parameter importance (EWC~\cite{kirkpatrick2017overcoming}) and distilling output~\cite{hinton2015distilling, yu2019learning} from old and new models (LwF~\cite{li2017learning}) to retain information about previous tasks. SDC~\cite{yu2020semantic} proposed compensating the feature drift to prevent forgetting.
Architecture-based methods assign specified parameters to tasks or dynamically expand the network architecture. HAT~\cite{serra2018overcoming} blocked and activated specific parts for old tasks, while DER~\cite{yan2021dynamically} utilized task-specific feature extractors and tackles parameter growth through pruning. Transformer-based continual learning paradigms, like DyTox~\cite{douillard2022dytox}, utilized shared self-attention layers and task-specific tokens to achieve task-specialized embeddings. CMAE~\cite{zhai2023masked} proposed to apply Masked Autoencoders (MAEs) for continual learning and introduced a bilateral MAE framework that integrates learning from both image-level and embedding-level. FKSR~\cite{zhai2024fine} proposed to adaptively assign distillation loss weights by evaluating the relevance of each patch to the current task.

\subsection{Continual Learning with Foundation Models}
A recent emerging trend in the field of continual learning is to combine pre-trained vision transformers with parameter-efficient fine-tuning techniques to continuously adapt the model to a stream of incoming downstream tasks. Specifically, these techniques involve prompt tuning~\cite{lester2021power, jia2022visual}, prefix tuning~\cite{li2021prefix}, Adapter\cite{pfeiffer2021adapterfusion, wang2020k}, LoRA~\cite{hulora} , etc. The core of applying such a technique to continual learning is to construct additional learnable parameters or modules to instruct pre-trained representations and select appropriate prompts during inference time. L2P~\cite{wang2022learning} optimized the cosine similarity between query features and learnable keys, the most relevant prompts are selected and prepended to the token sequence both during training and inference. DualPrompt~\cite{wang2022dualprompt} further subdivided the prompts into general prompts sharing across all tasks and expert prompts picked and optimized in the same way as~\cite{wang2022learning}. CODA-Prompt~\cite{smith2023coda} proposed to reweight prompts through input-conditioned weights, facilitating end-to-end optimization of the query-key mechanism across tasks. ConvPromt~\cite{roy2024convolutional} proposed to generate layer-wise prompts from shared embeddings and integrated Large Language Model-driven task similarity analysis to dynamically allocate prompt resources.
Distinct from the aforementioned prompt-based approaches, another line of work aims to construct classifiers through the extraction of robust feature representations via pre-trained models. SLCA~\cite{zhang2023slca} fine-tuned the full pre-trained model and recommended using different learning rates for the representation layers and the classifier to address the forgetting problem in representation layers, they further incorporated prototype replay for post-hoc alignment of the classification layers to reduce bias in the classifier. RanPAC~\cite{mcdonnell2024ranpac} suggested employing a frozen random projection layer to project pre-trained features into a high-dimensional space, thereby improving linear separability, they additionally utilize an online LDA classifier to eliminate correlations between categories. EASE~\cite{zhou2024expandable} proposed to employ distinct adapters for each task to acquire task-specific features, and construct a unified classifier by synthesizing prototypes for old classes through semantic relevance. Considering the feature generalizability of large-scale vision-language pre-trained models, a series of recent works has emerged that either specifically design for or leverage these types of models~\cite{wang2023attriclip,frascaroli2024clip,yu2025language,zhang2024overcoming,huang2024class,jhaclap4clip,zhou2025learning}. SPU~\cite{zhang2024overcoming} introduced a selective parameter update strategy to mitigate generic knowledge forgetting in foundation models, by localizing to key layers and fine-tuning only a small portion of task-relevant parameters. CLAP4CLIP~\cite{jhaclap4clip} addressed uncertainties in multi-modal interactions by probabilistically modeling the visual-guided text feature space and incorporating task-specific probabilistic adapter modules. Furthermore, it leverages CLIP's rich pre-trained knowledge for weight initialization and distribution regularization to effectively alleviate catastrophic forgetting. RAPF~\cite{huang2024class} proposed to adaptively adjusting affected old class representations using textual features, combined with a decomposed parameter fusion technique to stabilize the adapter. PROOF~\cite{zhou2025learning} proposed to learn new knowledge by training task-specific expandable projection layers and freezing old ones to prevent forgetting, while also designing a cross-modal fusion module to better integrate visual and textual information.

\subsection{Pre-trained Vision-Language Model}
Large-scale vision-language pre-trained models like CLIP~\cite{radford2021learning} and ALIGN~\cite{jia2021scaling} have emerged as powerful tools for representation learning due to their high transferability. Numerous studies~\cite{lu2022prompt, zhou2022conditional, zhou2022learning} have focused on adapting these models to downstream tasks. For instance, CoOp~\cite{zhou2022learning} proposed vectorizing prompts, keeping the CLIP model fixed while optimizing only the prompt vectors. However, it was observed in\cite{zhou2022conditional} that prompts learned using this method exhibited poor generalization performance on unseen classes. To address this limitation, they introduced a meta-net to generate an input-conditional token for each image. Some works attempt to simultaneously optimize visual and textual prompts to improve the adaptability of vision-language models for downstream tasks. MaPLe~\cite{khattak2023maple} proposed to promote strong coupling between prompts of two different modalities to improve the consistency between visual and linguistic representations. PromptSRC~\cite{khattak2023self} designed a regularization framework when training models on downstream tasks to prevent overfitting, which enhances the generalization capacity of the model. Differing from prompt-based approaches, some works try to incorporate adapters with MLPs into the transformer architecture to capture task-specific information. CLIP-adapter~\cite{gao2024clip} introduced extra bottleneck layers for learning new features. While these approaches have demonstrated significant improvements on various tasks compared to the adapted results provided by CLIP~\cite{radford2021learning}, they all shared the common limitation of freezing all the pre-trained parameters. 

\begin{figure*}[t]
    \centering
    \includegraphics[width=0.95\textwidth]{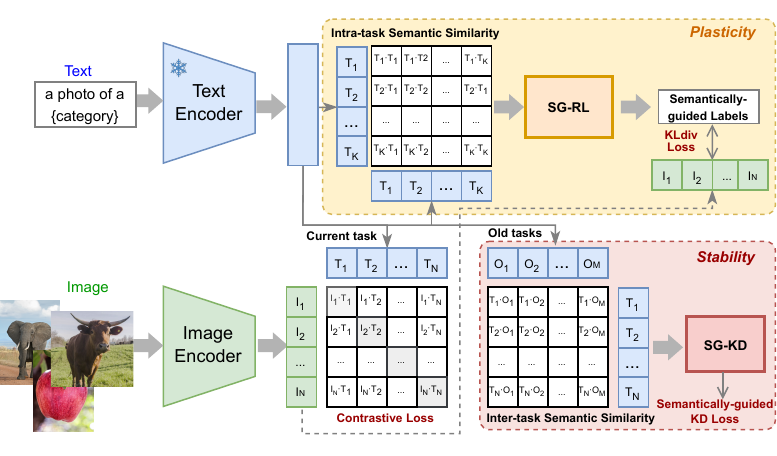} 
    
    \caption{An overview of the proposed framework. We update the image encoder part of the CLIP model and train it with contrastive loss. We integrate semantic information into two modules:
    The SG-RL module, represented by the yellow block of the framework, aims to learn more discriminative representation to improve the model plasticity based on the intra-task semantic similarity. We train the model with KL-divergence loss between the generated semantically-guided labels and the original predictions. The SG-KD module, represented by the pink block of the framework, exploits the semantic similarity between current and previous task labels for knowledge distillation,
    to consolidate the model stability.} 
    \label{fig:framework}
\end{figure*}

\section{Preliminary of CLIP model}
Here we explain the details of the CLIP model.
Let $I$ denote the set of normalized image features encoded by the image encoder, and $T$ denote the set of corresponding normalized text features. Assume that we have $N$ samples,
the logits for each image-text pair are computed as follows:
\begin{equation}\label{eq:p}
P_{i,j} = \beta\cdot I_{i} \cdot T_{j}^{\top},
\end{equation}
where $\beta$ is a learned scalar, $P_{i,j}$ is the cosine similarity score between the $i$th image and the $j$th text, and $i\in \{1, \dots, N\}$ and $j \in \{1, \dots, N\}$ are indices for the images and texts, respectively.
Then the loss for each modality can be expressed as:
\begin{align}
    L_{image} = -\frac{1}{N} \sum_{i=1}^N\log \left(\frac{\exp(P_{i,i})}{\sum_{j=1}^N \exp(P_{i,j})}\right),\\
    L_{text} = -\frac{1}{N} \sum_{j=1}^N\log \left(\frac{\exp(P_{j,j})}{\sum_{i=1}^N \exp(P_{i,j})}\right)
\end{align}

The overall loss of the CLIP model is the sum of the image loss $L_{image}$ and text loss $L_{text}$, which are averaged over the the number of samples. It can be represented as:
\begin{equation}
L_c= \frac{1}{2} (L_{image} + L_{text})
\label{eq:con_loss}
\end{equation}
The loss function is minimized during pre-training to learn the joint representations of images and texts where semantically similar pairs are closer together and dissimilar pairs are farther apart. 

The CLIP model has demonstrated remarkable continual learning performance without any fine-tuning~\cite{thengane2022clip}. To enhance its performance on new tasks and adapt to evolving data distributions over time, we thus propose to fine-tune the last block of the vision encoder.\footnote{We fine-tune different parts of the vision encoder in Table~\ref{tab:diff_layers}. Our experiments indicate that fine-tuning the last block of the encoder results in the best performance.} We aim to leverage the strong generalization ability of the CLIP model while also updating its representation to better adapt to incoming data.

\section{Method}
\label{sec:blind}

\begin{figure*}[t]
\centering
\begin{subfigure}[b]{0.42\textwidth}
    \raisebox{5pt}{\includegraphics[width=\textwidth]{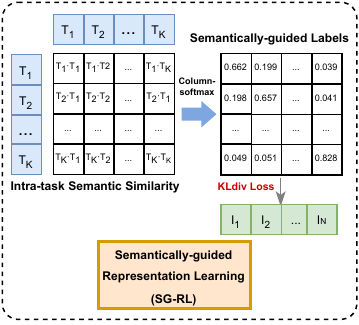}}
    \caption{Illustration of SG-RL module.}
    \label{fig:TLG}
  \end{subfigure}
  \hfill
  \begin{subfigure}[b]{0.545\textwidth}
    \includegraphics[width=\textwidth]{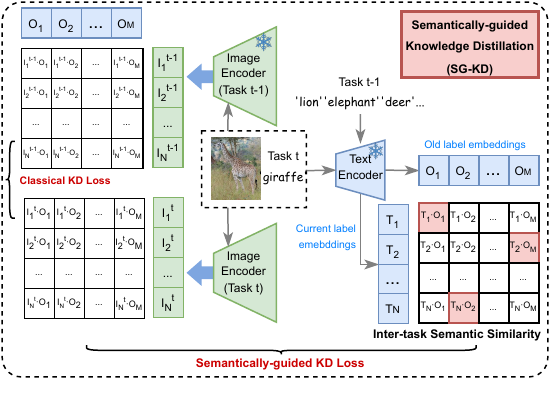}
    \caption{Illustration of SG-KD module.}
    \label{fig:TEKD}
  \end{subfigure}
  \caption{An illustration of two components. (a) The SG-RL module calculates the text embedding similarity between the current labels using a pre-trained text encoder. It then generates semantically-guided labels that are more informative. (b) The SG-KD module constructs an inter-task similarity matrix based on text embeddings of previous and current task class labels. By introducing semantic relationships between old and new classes without requiring any visual information, it enables the model to capture inter-task semantic connections and perform knowledge transfer, thereby enhancing stability.
  }
  \label{fig:detail}
\end{figure*}

To overcome the challenges of catastrophic forgetting during the training of pre-trained models, we introduce a novel approach that integrates \emph{semantic information} guidance into the process of continual knowledge learning, as shown in Fig.~\ref{fig:framework}. Leveraging the power of well-trained text embeddings, our proposed approach facilitates efficient interaction within and across task labels, leading to improved performance from two crucial perspectives. Firstly, we focus on learning more informative representations for new data by leveraging intra-task semantic similarity, thereby enhancing model plasticity (\emph{SG-RL} module). Secondly, we establish a relationship between old and new tasks by incorporating inter-task semantic similarity during the model distillation process, ensuring stability (\emph{SG-KD} module). 
Similar to recent methods~\cite{jhaclap4clip,zhou2025learning, zhang2024overcoming}, we also use a small set of exemplars in the form of old task images and text labels.
In the following sections, we introduce these two components and explain how they work.

%-------------------------------------------------------------------------
\subsection{Intra-task Semantically-Guided Representation Learning (SG-RL)}

As shown in the yellow part of Fig.~\ref{fig:framework}, assume we have $K$ classes contained in the current task $t$ (including exemplar classes, if applicable), their labels can be denoted as $C^t$=$\{c^t_1, c^t_2, \cdots, c^t_K\}$(for simplicity, we omit the superscript $t$ in the following descriptions). We encode the text embeddings of these labels by the pre-trained language part of CLIP
to obtain the normalized text embeddings $\mathbf{T}^{t}=\{T_1, T_2, \cdots, T_K\}$, where $||T_k||=1$ and $k\in \{1,2,\cdots,K\}$. After obtaining the text embeddings of the current labels, we acquire the \emph{intra-task semantic similarity} by computing the cosine similarity between each pair of text embeddings. This yields a text similarity matrix $\mathbf{S}^{t\leftrightarrow t}$  
which represents the intra-task similarity between the $i$-th and $j$-th labels; it can be described as follows:
\begin{equation}
    S_{i,j}^{t\leftrightarrow t}=[T_i^{\top} T_j]_{K\times K}, (i,j \in\{1, 2, \cdots,K\}),\label{eq:intra-sim}
\end{equation}
This similarity score contains the pairwise similarity within each category within the current task. 

To further convert the one-hot labels to our informative semantically-guided labels $C^{sg}=[C_{i,j}^{sg} ]_{K\times K} $, we compute the softmax function over each column of $\mathbf{S}^{t\leftrightarrow t}$ with a parameter $\alpha$ which controls the degree of softness of the generated labels as follows:
\begin{equation}
 C_{i,j}^{sg}= \frac{\exp(\alpha \cdot S_{i,j}^{t\leftrightarrow t})}{ {\textstyle \sum_{k=1}^{K}\exp (\alpha \cdot S_{i,k}^{t\leftrightarrow t})} } 
 \label{eq:sg_label}
\end{equation}
The semantically-guided labels $C^{sg}$ encode the relationships and similarities between categories, allowing for a more comprehensive representation of class associations within the current task. 

We then compute the KL-divergence loss $\mathcal{L}_{SG-RL}$ between $\hat{y}_{i,j}$ and the semantically-guided labels $C^{sg}_{i,j}$:
\begin{equation}
\mathcal{L}_{SG-RL}=D_{KL}(\hat{y} || C^{sg}) = \sum_{i=1}^{N}\sum_{j=1}^{K}\hat{y}_{i,j} \cdot \log\frac{\hat{y}_{i,j}}{C_{i,j}^{sg}},
\end{equation}
where the logits $P^t =[P_{i,j}^t]_{N\times K}$ of $N$ images in current task is computed by Eq.~\ref{eq:p},
and the prediction score $\hat{y}_{i,j}= \log \frac{\exp(P_{i,j}^t)}{\sum_{k=1}^{K}\exp(P_{i,k}^t)}$ is after applying the log-softmax.

The detailed illustration of the SG-RL model is shown in Fig.~\ref{fig:TLG}. We propose to exploit the semantic knowledge of the pre-trained text-encoder by means of the \emph{intra-task semantic similarity} between class labels (see Eq.~\ref{eq:intra-sim}). We replace the one-hot ground truth label in the cross entropy with a soft-assignment towards all current task classes. As a consequence, each sample contributes to aligning the vision backbone with respect to multiple semantic classes, leading to higher plasticity.

\subsection{Inter-task Semantically-Guided Knowledge Distillation (SG-KD)}
In this section, we aim to exploit the semantic knowledge of the pre-trained text-encoder to prevent forgetting and improve stability of the continual learner. Our main insight is that the semantic similarity between current and previous task labels can be exploited to prevent forgetting. For example, the current task label `truck' can be used to prevent forgetting of previous related class labels like `bus'. 
As shown Fig.~\ref{fig:TEKD}, given an image $x_i$ of the current task, the image features encoded by the previous image encoder are denoted as $I^{t-1}_{i}$, and the corresponding normalized text embeddings of old-class labels $y^{t-1}$
can be encoded by the pre-trained text encoder as $O_{j}$.  
Thus, the prediction logits $P^{t-1}=[P^{t-1}_{i,j}]_{N\times M}$ (the number of old classes is $M$) of current data on the previous image encoder is computed as:
\begin{equation}
P^{t-1}_{i,j} =\beta\cdot I^{t-1}_{i}\cdot O_{j}^{\top},  (i\in{1,2,\dots,N}, j\in{1,2,\dots,M})
\end{equation}\label{Eq:beta1}
The prediction logits of the current image encoder on old-class heads $P^t=[P^t_{i,j}]_{N\times M}$ are:
\begin{equation}
P^t_{i,j} = \beta\cdot I^t_{i} \cdot O_{j}^{\top},
\end{equation}
which can be used to compute a prediction over the labels:
\begin{equation}
    p^t(y^{t-1}_j|x_i) = \frac{\exp(P_{i,j}^{t})}{\sum_j \exp(P_{i,j}^{t})}
\end{equation}
and similarly for $p^{t-1}(y^{t-1}_j|x_i)$; the difference is that this probability is based on the model at time $t\text{-}1$. Traditional knowledge distillation~\cite{li2017learning,rebuffi2017icarl} between the old and current model can be described as follows:
\begin{equation}
    \mathcal{L}_{KD} = D_{KL}(p^t(y^{t-1}|x_i),p^{t-1}(y^{t-1}|x_i)).\label{eq:KD}
\end{equation}
Our contribution is that we aim to improve this distillation by also including the knowledge of the current sample labels about the previously seen classe in the distillation. 
Therefore, we use the semantic relationship between the current categories $y^{t}$ and the previous categories $y^{t-1}$, without requiring access to any visual information. We introduce Semantically-guided Knowledge Distillation (SG-KD); the loss for an image $x_i$ is given by: 
\begin{equation}
    \begin{split}
        \mathcal{L}_{SG-KD} = D_{KL}(p^t(y^{t-1}|x_i),p^{t-1}(y^{t-1}|x_i)) \\
         + \mu D_{KL}(p^t(y^{t-1}|x_i),s(y^{t-1}|y_{c_i}^t))\label{eq:sg-kd}
     \end{split}
\end{equation}
where $y^t_{c_i}$ refers to the image label of image $x_i$. The first term on the right-hand site is the same as in Eq.~\ref{eq:KD}. 
The second term provides us the distillation between $s(y^{t-1}|y_{c_i}^t)$ and the current predictions. $\mu$ is the tradeoff between these two distillation losses.
We derive $s(y^{t-1}|y^t_{c_i})$ from the \emph{inter-task semantic similarity} $\mathbf{S}^{{t-1}\leftrightarrow t}$ between the text embeddings of current task labels\footnote{The text embeddings of exemplar labels are integrated into $T_i$, when applicable.} $T$ and the old task labels $O$ as follows:
\begin{equation}
 S_{i,j}^{{t-1}\leftrightarrow t} = T_{i} \cdot O_{j}^{\top}
\end{equation}
Here $S_{i,j}^{{t-1}\leftrightarrow t}$ provides a measure of similarity between classes from different tasks. This allows the model to capture the semantic relationships between tasks and leverage this information for knowledge transfer.
Thus, $s(y^{t-1}|y^t_c)$ can be computed as:
\begin{equation} \label{eq:tau_s}
s(y^{t-1}_j|y^t_{c_i})=\frac{\exp(S_{i,j}^{{t-1}\leftrightarrow t}/\tau)}{\sum_j \exp(S_{i,j}^{{t-1}\leftrightarrow t}/\tau)}
\end{equation} 
$\tau$ is a hyper-parameter representing the temperature in the softmax function.

The final objective function can be described as follows:
\begin{equation}\label{eq:full}
    \mathcal{L}=\mathcal{L}_c+\lambda_1 \mathcal{L}_{SG-RL}+\lambda_2 \mathcal{L}_{SG-KD},
\end{equation}
where $\lambda_1$ and $\lambda_2$ are trade-offs between the contrastive loss $\mathcal{L}_c$, KL-divergence loss $\mathcal{L}_{SG-RL}$ and the semantically-guided distillation loss $\mathcal{L}_{SG-KD}$.

\section{Experiments}
\subsection{Experimental Settings}

\begin{table}[htb]
    \centering
    \caption{Details of all datasets used in this paper.}
\resizebox{\linewidth}{!}{
    \begin{tabular}{c|c|c|c}
\hline
Datasets            & \#Classes & Train size & Test size \\ \hline
CIFAR100~\cite{krizhevsky2009learning}           & 100       & 50000      & 10000     \\
ImageNet\_subset~\cite{deng2009imagenet}   & 100       & 129395     & 5000      \\
miniImageNet~\cite{deng2009imagenet}       & 100       & 50000      & 10000      \\
ImageNet1000~\cite{deng2009imagenet}      & 1000      & 1281167    & 50000     \\ 
Food-101~\cite{bossard2014food}           & 101       & 75750      & 25250     \\  
Stanford Cars~\cite{krause20133d}      & 196       & 8144       & 8041      \\ 
FGVC Aircraft~\cite{maji2013fine}      & 100       & 6667       & 3333      \\ 
Oxford-IIIT pets   & 37        & 3680       & 3669      \\ 
Caltech-101~\cite{CaltechDATA}        & 102       & 3060       & 6085      \\ 
Oxford Flowers 102~\cite{nilsback2008automated} & 102       & 2040       & 6149      \\  
CUB-200-2011~\cite{WahCUB_200_2011}       & 200       & 5994       & 5794      \\ 
Stanford Dogs~\cite{khosla2011novel}      & 120       & 12061      & 8519      \\ 
ImageNet-A ~\cite{hendrycks2021natural}      & 200       & 5981      & 1519      \\ 
ImageNet-R ~\cite{hendrycks2021many}      & 200       & 24000      & 6000      \\ 
VTAB ~\cite{zhou2025revisiting}      & 50       & 1796      & 8619      \\ 
 \hline
\end{tabular} 
}
\label{tab:all_datasets}
\end{table}

\begin{table*}[t]
\centering
\caption{The results on CIFAR100 
under 10, 20, and 50 steps. 
The bold parts represent the best results, underlined parts indicate the second best.
PTM-based methods were re-implemented with the same CLIP ViT-B/16 weights for fair comparison. We denote if the methods could be used for zero-shot classification (ZS) which is true (\Checkmark) for CLIP-based methods and false (\XSolidBrush) for methods using only vision backbones. We also denote the methods with learnable backbone (LB) by (\Checkmark) and the prompt- or adapter-based methods having frozen backbone by (\XSolidBrush).
}
\resizebox{0.8\linewidth}{!}{
\begin{tabular}{c|c|c|cc|cc|cc}
\hline
CIFAR100    &    &   & \multicolumn{2}{c|}{\textbf{10 steps}}         & \multicolumn{2}{c|}{\textbf{20 steps}}              & \multicolumn{2}{c}{\textbf{50 steps}}                         \\ \hline
Methods     & ZS  & LB  & \multicolumn{1}{c}{Avg Acc.}   & Last Acc.               & \multicolumn{1}{c}{Avg Acc.}        & Last Acc.             & \multicolumn{1}{c}{Avg Acc.}        & Last Acc.       \\ \hline
\rule{0pt}{2ex} 
Linear probe  & -  & - & \multicolumn{1}{c|}{-}    & 83.1               & \multicolumn{1}{c|}{-}         & 83.1               & \multicolumn{1}{c|}{-}          & 83.1       \\
Joint         & -  & - & \multicolumn{1}{c|}{-}    & 89.3               & \multicolumn{1}{c|}{-}         & 89.3               & \multicolumn{1}{c|}{-}          & 89.3       \\ \hline
\rule{0pt}{2ex} 

DER w/o p\cite{yan2021dynamically} & \XSolidBrush   & \Checkmark          & \multicolumn{1}{c|}{75.4} & 65.2       & \multicolumn{1}{c|}{74.1} & 62.5      & \multicolumn{1}{c|}{72.4} & 59.1   \\ 
DER\cite{yan2021dynamically}       & \XSolidBrush      & \Checkmark       & \multicolumn{1}{c|}{74.6} & 64.4       & \multicolumn{1}{c|}{74.0} & 62.6      & \multicolumn{1}{c|}{72.1} & 59.8   \\ 
DyTox++\cite{douillard2022dytox}   & \XSolidBrush   & \Checkmark          & \multicolumn{1}{c|}{77.0} & 67.5       & \multicolumn{1}{c|}{76.8} & 64.3      & \multicolumn{1}{c|}{75.5} & 59.5   \\ \hdashline

L2P \cite{wang2022learning}         & \XSolidBrush     & \XSolidBrush        & \multicolumn{1}{c|}{61.4} &57.8        & \multicolumn{1}{c|}{63.2} & 60.9      & \multicolumn{1}{c|}{61.7} & 61.8 \\
DualPrompt \cite{wang2022dualprompt}& \XSolidBrush     & \XSolidBrush        & \multicolumn{1}{c|}{76.8} &67.4        & \multicolumn{1}{c|}{72.9} & 66.4      & \multicolumn{1}{c|}{73.0} & 70.3 \\
RanPAC \cite{mcdonnell2024ranpac}  & \XSolidBrush  & \XSolidBrush  & \multicolumn{1}{c|}{\underline{88.6}}  &\underline{82.0}  & \multicolumn{1}{c|}{\underline{88.9}}    & \underline{82.0}  & \multicolumn{1}{c|}{\underline{89.0}}          & \underline{81.8}  \\
CODA-Prompt \cite{smith2023coda}   & \XSolidBrush      & \XSolidBrush        & \multicolumn{1}{c|}{78.1} & 67.8       & \multicolumn{1}{c|}{71.4} & 60.3      & \multicolumn{1}{c|}{47.8}   & 36.0 \\
ConvPrompt  \cite{roy2024convolutional} & \XSolidBrush   & \XSolidBrush      & \multicolumn{1}{c|}{80.1} & 70.3       & \multicolumn{1}{c|}{73.7} & 62.5      & \multicolumn{1}{c|}{53.2}   & 40.1 \\
EASE   \cite{zhou2024expandable}     & \XSolidBrush     & \XSolidBrush       & \multicolumn{1}{c|}{82.4} & 73.4       & \multicolumn{1}{c|}{79.8} & 67.9      & \multicolumn{1}{c|}{53.8}   & 32.7 \\ \hdashline
\rule{0pt}{2ex} 
Continual-CLIP\cite{thengane2022clip}  & \Checkmark      & \XSolidBrush      & \multicolumn{1}{c|}{-}    & 68.7       & \multicolumn{1}{c|}{-}     & 68.7      & \multicolumn{1}{c|}{-}          & 68.7  \\
CoOp \cite{zhou2022learning}     & \Checkmark     & \XSolidBrush        & \multicolumn{1}{c|}{74.9} & 66.6       & \multicolumn{1}{c|}{75.1}  & 65.9      & \multicolumn{1}{c|}{79.2}   & 69.9 \\
CoCoOp \cite{zhou2022conditional}& \Checkmark     & \XSolidBrush        & \multicolumn{1}{c|}{71.6} & 64.1       & \multicolumn{1}{c|}{77.3}  & 67.5      & \multicolumn{1}{c|}{80.5}   & 71.5 \\
SPU   \cite{zhang2024overcoming} & \Checkmark     & \Checkmark          & \multicolumn{1}{c|}{87.6} & 80.0       & \multicolumn{1}{c|}{87.4}  & 80.5      & \multicolumn{1}{c|}{86.9}   & 78.8 \\
CLAP4CLIP  \cite{jhaclap4clip}   & \Checkmark     & \XSolidBrush        & \multicolumn{1}{c|}{86.5} & 78.6       & \multicolumn{1}{c|}{86.0}  & 77.4      & \multicolumn{1}{c|}{82.6}   & 75.0 \\
RAPF   \cite{huang2024class}     & \Checkmark     & \XSolidBrush        & \multicolumn{1}{c|}{86.4} & 79.6       & \multicolumn{1}{c|}{86.0}  & 78.0      & \multicolumn{1}{c|}{81.3}   & 69.9 \\
PROOF  \cite{zhou2025learning}   & \Checkmark     & \XSolidBrush        & \multicolumn{1}{c|}{83.0} & 73.8       & \multicolumn{1}{c|}{83.4}  & 73.6      & \multicolumn{1}{c|}{83.5}   & 72.7 \\
\rowcolor{LightPurple}
Ours            & \Checkmark   & \Checkmark   & \multicolumn{1}{c|}{\textbf{89.6}}       & \textbf{83.3}    & \multicolumn{1}{c|}{\textbf{89.6}}           &\textbf{82.8}        & \multicolumn{1}{c|}{\textbf{89.5}}       &\textbf{82.2}            \\ \hline
\end{tabular}
}
\label{tab:cifar_3}
\end{table*}

\begin{table*}[t]
\centering
\caption{We compare the results on the ImageNet\_subset dataset (100 classes) and the ImageNet1000 full dataset (1000 classes), where the training session is split into 10 steps, with each step containing an equal number of classes for each task. }
\resizebox{0.8\linewidth}{!}{
\begin{tabular}{c|c|c|cccc|cccc}
\hline
\multirow{3}{*}{Methods}   & &  & \multicolumn{4}{c|}{\textbf{ImageNet\_subset 10 steps}}                     & \multicolumn{4}{c}{\textbf{ImageNet1000 10 steps}}             \\ \cline{2-11} 
                           & &   & \multicolumn{2}{c}{top-1}  & \multicolumn{2}{c|}{top-5}                     & \multicolumn{2}{c}{top-1}   & \multicolumn{2}{c}{top-5}        \\ \cline{2-5} \cline{6-11}  & ZS  & LB  
                             & \multicolumn{1}{c}{Avg}& Last  & \multicolumn{1}{c}{Avg}   & Last         & \multicolumn{1}{c}{Avg}   & Last  & \multicolumn{1}{c}{Avg}   & Last \\ \hline
Linear probe         &-&-        & -     & \multicolumn{1}{c|}{83.9}  &- &97.6                                      & -     & \multicolumn{1}{c|}{80.2}    &- &94.1 \\  
Joint              &-&-          & -     & \multicolumn{1}{c|}{86.1}  &- &98.3                                      & -     & \multicolumn{1}{c|}{81.1}    &- &96.5   \\ \hline
DER w/o p\cite{yan2021dynamically}  & \XSolidBrush   & \Checkmark    & 77.2 & \multicolumn{1}{c|}{66.7} &93.2 &87.5                         & 68.8  & \multicolumn{1}{c|}{60.2} &88.2 &82.9\\ 
DER\cite{yan2021dynamically}       & \XSolidBrush   & \Checkmark       & 76.1 & \multicolumn{1}{c|}{66.1} &92.8 &88.4                         & 66.7  & \multicolumn{1}{c|}{58.6} &87.1 &81.9 \\ 
DyTox++\cite{douillard2022dytox}    & \XSolidBrush   & \Checkmark      & 80.8 & \multicolumn{1}{c|}{72.5} &94.4 &90.1                         & 73.2  & \multicolumn{1}{c|}{64.6} &91.1 &87.1  \\  \hdashline
L2P \cite{wang2022learning}     & \XSolidBrush   & \XSolidBrush          & 68.7 & \multicolumn{1}{c|}{59.7} &91.1 &85.4                         & 64.6  & \multicolumn{1}{c|}{58.8} &88.6 &81.1\\
DualPrompt \cite{wang2022dualprompt}    & \XSolidBrush   & \XSolidBrush    & 81.7 & \multicolumn{1}{c|}{71.3} &98.5 &96.9                         & 73.4  & \multicolumn{1}{c|}{65.6} &95.6 &93.2\\
RanPAC    \cite{mcdonnell2024ranpac}    & \XSolidBrush   & \XSolidBrush    & \textbf{90.7} & \multicolumn{1}{c|}{\textbf{83.7}} &98.3 &96.8       &\underline{81.8} &\multicolumn{1}{c|}{\underline{73.9}} &96.2 &93.7\\
CODA-Prompt   \cite{smith2023coda}    & \XSolidBrush   & \XSolidBrush      & 81.1 & \multicolumn{1}{c|}{66.8} &97.8 &95.3                         &81.7   &\multicolumn{1}{c|}{73.6}  &\underline{97.0} &\underline{94.9}\\
ConvPrompt   \cite{roy2024convolutional}  & \XSolidBrush   & \XSolidBrush   & 73.3 & \multicolumn{1}{c|}{65.0} &94.1 &88.1                         &76.2   &\multicolumn{1}{c|}{70.8}  &94.5 &92.2\\
EASE   \cite{zhou2024expandable}     & \XSolidBrush   & \XSolidBrush       & 83.6 & \multicolumn{1}{c|}{71.0} &98.5 &96.6                         &79.1   &\multicolumn{1}{c|}{70.3}  &96.3 &93.7\\ \hdashline
Continual-CLIP\cite{thengane2022clip}  & \Checkmark     & \XSolidBrush  & -    & \multicolumn{1}{c|}{75.2} &-    &96.9                         & -     & \multicolumn{1}{c|}{68.6} &- &90.6\\ 
CoOp \cite{zhou2022learning}  & \Checkmark     & \XSolidBrush             & 82.2 & \multicolumn{1}{c|}{72.0} &98.8 &\underline{97.6}             &75.1   & \multicolumn{1}{c|}{67.5} &96.3 &89.7\\
CoCoOp \cite{zhou2022conditional} & \Checkmark     & \XSolidBrush         & 80.1 & \multicolumn{1}{c|}{69.1} &97.9 &95.4                         &69.2   & \multicolumn{1}{c|}{58.9} &89.4 &82.3\\ 
SPU \cite{zhang2024overcoming}  & \Checkmark     & \Checkmark          & 86.5 & \multicolumn{1}{c|}{77.6} &97.6 &95.2                         &78.7   &\multicolumn{1}{c|}{70.7}  &94.1 &91.0\\
CLAP4CLIP \cite{jhaclap4clip}  & \Checkmark     & \XSolidBrush           & 87.0 & \multicolumn{1}{c|}{79.1} &96.6 &93.8                         &81.5   &\multicolumn{1}{c|}{74.0}  &94.7 &91.5\\
RAPF \cite{huang2024class}  & \Checkmark     & \XSolidBrush               & 88.3 & \multicolumn{1}{c|}{80.2} &\underline{98.9} &97.5             &79.9   &\multicolumn{1}{c|}{67.3}  &95.1 &89.5\\
PROOF \cite{zhou2025learning} & \Checkmark     & \XSolidBrush            & 82.3 & \multicolumn{1}{c|}{65.6} &98.1 &95.4                         &70.1   &\multicolumn{1}{c|}{56.5}  &92.9 &87.2\\

\rowcolor{LightPurple}
Ours   & \Checkmark     & \Checkmark          & \underline{89.8}   & \multicolumn{1}{c|}{\underline{83.1}}  &\textbf{99.1} &\textbf{98.4}  & \textbf{83.4} & \multicolumn{1}{c|}{\textbf{75.1}}  &\textbf{97.4} &\textbf{95.2}   \\ \hline
\end{tabular}}
\label{tab:sub_1k}
\end{table*}

\subsubsection{Dataset and task spilt.}\label{sec:split} Our experimental evaluation begins by conducting experiments on three datasets that are commonly used for continual learning scenarios: CIFAR100~\cite{krizhevsky2009learning}, ImageNet\_subset, and ImageNet1000 ~\cite{deng2009imagenet}. For CIFAR100, we adopt the incremental phase splitting into 10, 20, and 50 steps, following the approach of DyTox~\cite{douillard2022dytox}. 
Regarding ImageNet\_subset and ImageNet1000, we split the incremental phase into 10 steps, with 10/100 new classes added at each incremental step. We further evaluate the performance on eight fine-grained datasets. 
We also evaluate for few-shot continual learning on miniImageNet, and CUB-200-2011 following~\cite{zhang2021few,zhou2022forward}. For miniImageNet, we split the 100 classes into 60 base classes and 40 classes across 8 sessions with 5 new classes per session. CUB-200-2011 is divided into 100 base classes and 100 classes across 10 sessions with 10 new classes per session. Each incremental session for all three datasets consists of 5 training samples per new class. 

The dataset splits follow two patterns: in the $A + B \times C $ split, $A$ represents the number of classes in the initial task, $C$ indicates the total number of steps, and $B$ signifies the number of new classes added at each incremental stage; for the $A \times B$ split, $A$ new classes are introduced in each of the $B$ incremental steps. Note that we handle the Oxford-IIIT pets dataset separately due to its 12 cat categories and 25 dog categories, splitting it into two stages: one for cats and another for dogs. The details of all datasets are listed in Table~\ref{tab:all_datasets}.

\subsubsection{Implementation details.}

We train every task for 10 epochs except for CUB and Aircrafts with 20 epochs. We adopt SGD in all experiments as the optimizer with an initial learning rate of 0.01(0.001 for few-shot setting), weight decay is 2e-4 and 0.9 for momentum. Batch size is 256 for all experiments. We set $\alpha$ to 13 in Eq.~\ref{eq:sg_label}, $\beta$ is set to 100 in Eq.~\ref{Eq:beta1} following~\cite{thengane2022clip} and $\tau$ to 0.1 in Eq.~\ref{eq:tau_s} for all datasets. The trade-off parameters, $\lambda_1$ and $\lambda_2$, are set to 0.5 and 0.1 across all datasets (the hyper-parameters sensitive experiments can be found in Section~\ref{section:hyper} ). We save 20 exemplars for each old class by herding algorithm in all experiments except the ablation on different exemplar sizes (in Section~\ref{sm:emexplar}), following the setting in~\cite{douillard2022dytox, rebuffi2017icarl, yan2021dynamically}. Under the few-shot setting, none exemplar is saved following the setting in~\cite{zhang2021few, zhou2022forward}. For all PTM based methods, we re-implemented with the same pre-trained ViT-B/16 CLIP~\cite{radford2021learning} weights for fair comparison.\footnote{ For Continual-CLIP~\cite{thengane2022clip}, since the exact prompt and specific class names employed by CLIP~\cite{radford2021learning} for each dataset are unknown. We tried several prompts and finally for each dataset we utilize a specific prompt so that the results of our own implementation are comparable or identical to those provided by CLIP~\cite{radford2021learning}.}.
For implementing `Linear Probe', we directly add a fully connected layer after the linear projection, we train 100 epochs for each dataset with an initial learning rate of 0.1 and decays by 0.1 every 45 epochs. For `Joint', 
all training samples are acquired concurrently in the same session. We train 100 epochs for each dataset with an initial learning rate of 0.01 and decays by 0.1 every 45 epochs.

For CIFAR100, we refined this general configuration to ensure fair comparison with recent highly-optimized methods~\cite{zhang2024overcoming}, given its unique low-resolution nature.
Specifically, for CIFAR100 only, we activated all attention projection matrices in residual blocks from shallow to deep layers to better capture low-level features, we set the learning rate to 5e-6 and adopted the AdamW optimizer.

For all re-implementation details of compared methods, please refer to Appendix~\ref{re_details}.

\subsubsection{Evaluation metrics}
Following the evaluation protocol of previous works~\cite{douillard2022dytox,wang2022learning,zhang2023slca}, we report the average incremental accuracy `Avg' and the final accuracy `Last'. For a sequence of tasks denoted as $\mathcal{T} = \{1, 2, \dots, n\}$, let $A_{t}$ represent the mean accuracy on all learned categories after learning task $t$, then `Avg' can be denoted as $Avg= \frac{1}{n} {\textstyle \sum_{1}^{n}}A_{t}$, and the final accuracy `Last' is $A_{n}$ which represent the average accuracy of all categories after learning the last task $n$. Unless explicitly stated, all experiments and ablation studies herein will report only the final accuracy `Last'.

\begin{table*}[t]
\centering
\caption{Last accuracy on eight fine-grained datasets with different task splits.
}
\resizebox{\linewidth}{!}{
\begin{tabular}{c|c|c|c|c|c|c|c|c|c|c}
\hline
\textbf{Dataset}       & \textbf{ZS} & \textbf{LB}                & \textbf{Food}    & \textbf{Cars}   & \textbf{Aircraft}  & \textbf{Pets}  & \textbf{Caltech}   & \textbf{Flowers}  & \textbf{CUB}   & \textbf{Dogs} \\ \hline
\#classes              & - & -                & 101     & 196    & 100      & 37     & 102      & 102     & 200    & 120                                                                           \\ 
split              & - & -                    & 11+$9 \times 10$ & $28 \times 7$   & $10 \times 10$    & 12+25  & $17 \times 6$     & $17\times6$   & $20\times10$  & $12\times10$                 \\ \hline 
Linear probe     & - & -                      & 92.8    & 86.7   & 59.5     & 93.1   & 94.7     & 98.1 & 80.4$^*$  & 78.6$^*$                                                                    \\
Joint      & - & -                            & 93.1    & 91.5   & 73.2     & 94.7   & 93.8     & 98.1  & 85.0  & 80.6  \\ \hline 
L2P \cite{wang2022learning}      & \XSolidBrush   & \XSolidBrush         & 74.9    & 63.6   & 27.3     & 83.3   & 73.1     & 77.6  & 70.0  & 58.9    \\
DualPrompt \cite{wang2022dualprompt}  & \XSolidBrush   & \XSolidBrush    & 85.2    & 80.9   & 49.9     & 90.5   & 90.5     & \underline{96.8}  & 76.2  & 69.8    \\

RanPAC \cite{mcdonnell2024ranpac}    & \XSolidBrush   & \XSolidBrush    & \textbf{92.0}    & \underline{86.3} & \underline{62.3}  & 91.7             & \textbf{94.6}    & \textbf{97.7}    & \underline{82.7} & \underline{77.0}   \\
CODA-Prompt   \cite{smith2023coda}   & \XSolidBrush   & \XSolidBrush       & 83.1             & 72.2             & 27.7              & 90.1             & 85.6             & 88.3             & 63.6             & 58.4   \\
ConvPrompt   \cite{roy2024convolutional}  & \XSolidBrush   & \XSolidBrush  & 84.1             & 86.3             & 53.7              & 91.1             & 88.4             & 95.4             & 79.1             & 71.1   \\
EASE   \cite{zhou2024expandable}    & \XSolidBrush   & \XSolidBrush        & 86.5             & 76.7             & 41.3              & 92.6             & 91.8             & 93.5             & 73.0             & 69.8   \\ \hdashline
Continual-CLIP\cite{thengane2022clip} & \Checkmark     & \XSolidBrush  & 89.2             & 65.6             & 27.1              & 88.9             & 89.3             & 70.4              & 55.6             & 63.4   \\
CoOp \cite{zhou2022learning}   & \Checkmark     & \XSolidBrush         & 84.7             & 81.7             & 48.4              & 93.3             & 86.9             & 94.2              & 78.1             & 74.0   \\
CoCoOp \cite{zhou2022conditional}   & \Checkmark     & \XSolidBrush    & 86.0             & 74.9             & 39.4              & 93.2             & 86.6             & 93.0              & 70.0             & 69.7   \\
SPU   \cite{zhang2024overcoming}   & \Checkmark     & \Checkmark     & 77.2             & 68.5             & 59.7              & \underline{93.1} & 69.6             & 90.0              & 66.3             & 67.4   \\
CLAP4CLIP   \cite{jhaclap4clip}   & \Checkmark     & \XSolidBrush      & 74.6             & 72.4             & 45.4              & 91.4             & 91.3             & 93.7              & 82.0             & 61.8   \\
RAPF   \cite{huang2024class}      & \Checkmark     & \XSolidBrush      & 83.0             & 44.1             & 31.0              & 79.0             & 66.8             & 70.2              & 27.0             & 47.8   \\
PROOF   \cite{zhou2025learning}   & \Checkmark     & \XSolidBrush      & 84.9             & 80.0             & 51.0              & 89.7             & 90.4             & 93.8              & 74.1             & 69.9   \\
\rowcolor{LightPurple}
Ours  & \Checkmark     & \Checkmark  & \underline{90.9} & \textbf{88.6}    & \textbf{66.6}     & \textbf{94.0}    & \underline{92.3} & 96.2             & \textbf{83.7}    & \textbf{79.6}    \\ \hline
\end{tabular}}
\label{tab:fg_dataset}
\end{table*}

\subsection{Full-shot Continual Learning Setting}

\subsubsection{Evaluation on general datasets.} 
We conducted extensive experiments on three benchmark datasets, namely CIFAR100, ImageNet\_subset, and ImageNet1000 in comparison to state-of-the-art approaches. In Table~\ref{tab:cifar_3}, we present the averaged accuracy on CIFAR100 over three different class orders and three different splits. We categorize all compared methods into three groups: (1) training from scratch, (2) training with a vision-only backbone, and (3) training with a vision-language backbone. Additionally, we indicate whether each method supports zero-shot (ZS) classification. This capability is marked as (\Checkmark{}) for CLIP-based approaches and (\XSolidBrush{}) for methods relying solely on vision backbones. We also denote the methods with learnable backbone (LB) by (\Checkmark) and the prompt- or adapter-based methods having frozen backbone by (\XSolidBrush)\footnote{Fine-tuning results for the frozen backbones are reported in Appendix~\ref{LB}.}. Under the standard 10-step class-incremental split, our method achieves a final accuracy of 83.3\%, outperforming the previous state-of-the-art by 1.3\% points. Furthermore, for more challenging incremental sequences of 20 and 50 steps, our approach consistently surpasses RanPAC, the strongest existing baseline in terms of the \emph{`Last'} accuracy. It is worth noting that RanPAC operates in a high-dimensional space ($d=10000$), in contrast to most other methods which use default embedding dimensions for classification. 
Moving to larger-scale continual learning scenarios (Table~\ref{tab:sub_1k}), our method also demonstrates strong performance. On the ImageNet\_subset dataset, it achieves superior top-5 Last accuracy, surpassing the performance of linear probing~\cite{radford2021learning}, and yields comparable top-1 Last accuracy. On the more challenging ImageNet1000, our method outperforms the state-of-the-art approach by 1.2\% in top-1 Last accuracy, while also achieving superior results in top-5 accuracy.
Overall, our experimental results demonstrate that the proposed method consistently outperforms state-of-the-art approaches on all three datasets, showcasing its superior performance and stability in continual learning settings.

\begin{table*}[t]
\centering
\caption{Last accuracy on ImageNet-A ($20 \times 10$), ImageNet-R ($20 \times 10$) and VTAB ($10 \times 5$) .}
\resizebox{\textwidth}{!}{
\begin{tabular}{cccccccccc}
\toprule
\textbf{Method}   & RanPAC & CODA-Prompt & ConvPrompt & EASE & SPU & CLAP4CLIP & RAPF & PROOF & Ours \\ \midrule
ImageNet-A        & 57.9   & 43.8        & 55.6       & 51.9 & 55.4 & 45.6 & 38.5 & 45.1  & \textbf{58.7} \\
ImageNet-R        & 78.0   & 74.7        & 78.6       & 76.8 & 79.6 & 80.7 & 78.7 & 74.7  & \textbf{82.1} \\
VTAB              & 90.2   & 78.9        & 61.9       & 89.2 & 93.0 & 91.5 & 81.7 & 72.5  & \textbf{94.0} \\
\bottomrule
\end{tabular}}
\label{tab:large_gap}
\end{table*}

\begin{figure*}[t]
    \centering
    \includegraphics[width=0.8\textwidth]{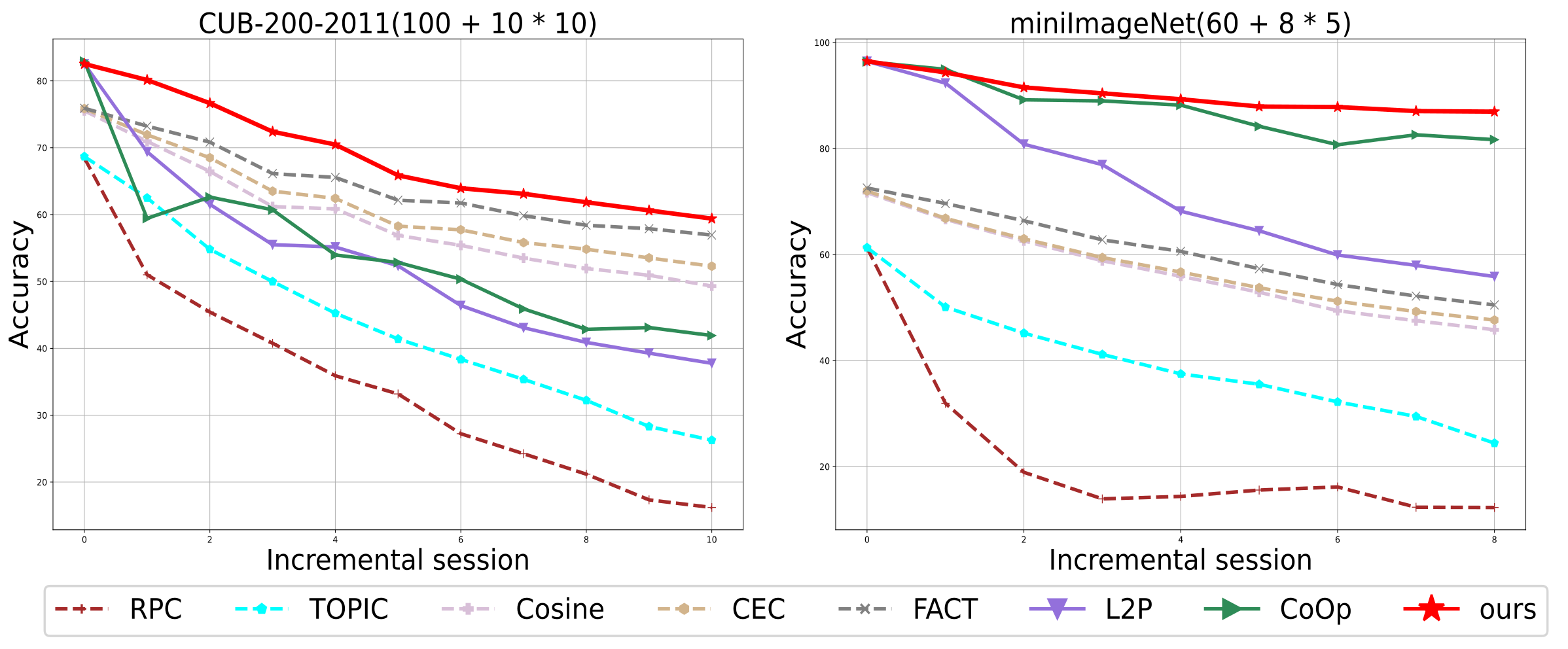}
    \caption{The accuracy change during few-shot incremental training sessions on CUB-200-2011 and miniImageNet datasets. } 
    \label{fig:few_shot_dynamic}
\end{figure*}

\begin{table*}[t]
\centering
\caption{The effect of different components. \emph{`FT'} represents only applying contrastive loss for training, \emph{`One-hot Label'} utilizes the prompt of each class as a classification head during training, and uses the one-hot encoded labels to calculate the cross-entropy loss. \emph{`Naive KD'} distills the output of the current data between old and new models with the KD loss proposed in\cite{hinton2015distilling} (ref Eq.~\ref{eq:KD}). }
\resizebox{\linewidth}{!}{
\begin{tabular}{c|c|c|c|c|c|c|c}
\hline
          & FT        & One-hot Label   & SG-RL(Eq.~\ref{eq:sg_label})   & Naive KD(Eq.~\ref{eq:KD})   & SG-KD(Eq.~\ref{eq:sg-kd})  & \textbf{CIFAR100 10 steps} &\textbf{ImageNet\_subset 10 steps} \\ \hline
C-CLIP\cite{thengane2022clip}  &              &             &            &         &            & 68.7         & 75.2      \\ \hline
          & \checkmark         &              &             &            &                      & 80.1         & 81.1      \\ 
          & \checkmark         & \checkmark   &             &            &                      & 80.5         & 75.2      \\ 
          & \checkmark         &              & \checkmark  &            &                      & 81.9         & 81.8      \\ 
          & \checkmark         &              & \checkmark  & \checkmark &                      & 82.4         & 82.6      \\ 
          & \checkmark         &              & \checkmark  &            & \checkmark           & 83.3         & 83.1      \\ \hline
\end{tabular}}
\label{tab:ab_com}
\end{table*}

\subsubsection{Evaluation on fine-grained datasets.}
To further validate the effectiveness of the proposed method, we conduct experiments on eight fine-grained datasets. In fine-grained datasets, all categories belong to the same parent category in a hierarchical relationship, which intuitively indicates a high degree of semantic similarity among them, which is more challenging. In Table~\ref{tab:fg_dataset}, we present the performance of our proposed method compared to other methods on different splits of the datasets. 
Our method achieves the best performance on five datasets, with improvements of up to 4.3\%, and delivers competitive results on the remaining benchmarks.
Moreover, our method even surpasses the results of linear probing evaluation on multiple datasets, which is typically used to assess the quality of pre-trained features.

The results of our method combined with the RanPAC classifier are presented in Appendix~\ref{comb}, further validating the strength of the learned feature representations.

\subsection{Generalization Performance with Semantic Gaps}
We have conducted experiments on three challenging datasets known for their significant domain shift from pre-training data: ImageNet-A~\cite{hendrycks2021natural}, ImageNet-R~\cite{hendrycks2021many} and VTAB~\cite{zhai2019large, zhou2025revisiting}. ImageNet-A primarily consists of `natural adversarial examples' or hard samples where models often struggle. ImageNet-R contains renditions of ImageNet classes in various artistic styles (e.g., paintings, sketches, cartoons), presenting a substantial stylistic shift. VTAB~\cite{zhou2025revisiting} was collected from 5 datasets with distinctly different distributions, including: the RESISC45 remote sensing image scene classification dataset, the DTD texture description dataset, the Pets dataset, the EuroSAT satellite imagery dataset, and the Flowers dataset. Each dataset is assigned to a task and contains 10 classes. These datasets are widely used in recent class-incremental learning research to evaluate model performance when faced with a significant domain gap between pre-train and downstream data~\cite{zhou2025revisiting, zhou2024expandable, zhou2024continual}.

The results of the experiments are in Table~\ref{tab:large_gap}. These results demonstrate that our method achieves competitive/superior performance on all datasets, indicating its robust generalization capability even when faced with considerable semantic and stylistic gaps between the pre-training and downstream data. Compared to all added baselines, our method shows better resilience to such domain shifts.

\subsection{Few-shot Continual Learning Setting}

Our proposed method has demonstrated impressive results on several datasets, including coarse and fine-grained datasets. It is interesting to evaluate its effectiveness under the few-shot continual learning setting, which poses a significant challenge. In this setting, the model needs to quickly adapt to new classes with limited labeled data while preserving knowledge of previous tasks. We present dynamic accuracy curves that showcase the incremental training sessions on the CUB-200-2011 and miniImageNet datasets in Fig.~\ref{fig:few_shot_dynamic}. The curves highlight the notable performance of our method (shown in red) as it significantly outperforms state-of-the-art approaches with (in solid line) and without (in dashed line) ViT backbone by a substantial margin over all incremental sessions. Particularly on the miniImageNet dataset, our method achieves a final accuracy about 36.5 points higher than FACT and 5.3 points higher than CoOp.

Additionally, we evaluate our method alongside several baselines to assess their ability to preserve zero-shot performance in multi-domain task-incremental learning. Detailed results are presented in Appendix~\ref{MTIL}.

\subsection{Ablation Study}

\subsubsection{Components of our proposed method}
We conduct ablation experiments on CIFAR100 and ImageNet\_subset to thoroughly examine the effect of the different proposed modules. Both datasets were split into 10 steps, and the results are shown in Table~\ref{tab:ab_com}. On both datasets, the contrastive loss significantly improves the final accuracy compared to Continual-CLIP. However, when we additionally apply the one-hot encoded labels to add an auxiliary classification loss, we observe a significant performance drop. By incorporating our \emph{`SG-RL'} module, the model shows a gain of 1.8 
points on CIFAR100 and 0.7 points on ImageNet\_subset. Furthermore, when we apply knowledge distillation to the loss function, our proposed \emph{`SG-KD'} demonstrates superior performance over \emph{`Naive KD'} (ref Eq.~\ref{eq:KD}) on both datasets, with a 
gain of 0.9
points on CIFAR100. 

\begin{table}[t]
\centering
\caption{Comparison of different vision models on ImageNet\_subset, learned with 10 steps. }
\resizebox{0.8\linewidth}{!}{
\begin{tabular}{c|c|cc}
\hline
\multirow{2}{*}{}         &                                 & \multicolumn{2}{c}{ImageNet\_subset} \\ \cline{1-4}
                          &                                 & Avg        & Last       \\ \hline
\multirow{2}{*}{ResNet 50}& C-CLIP\cite{thengane2022clip}   & -          & 65.6                \\
                          & Ours                            & 80.2       & 71.3       \\ \hline
\multirow{2}{*}{ViT-B/16} & C-CLIP\cite{thengane2022clip}   & -          & 75.2          \\
                          & Ours                            & 89.8       & 83.1         \\ \hline
\multirow{2}{*}{ViT-L/14} & C-CLIP\cite{thengane2022clip}   & -          & 81.5        \\
                          & Ours                            & 91.9       & 85.8       \\ \hline
\end{tabular}}
\label{tab:diff_backbones}
\end{table}

\begin{table*}[t]
\centering
\caption{Comparison of training different layers of image encoder.}
\resizebox{0.9\linewidth}{!}{
\begin{tabular}{c|c|c|c|c|c|c}
\hline
 &  & Last 6 blocks & Last 3 blocks & Proj. + last block & Proj. & \textbf{Last block} \\ \hline
\multirow{2}{*}{ImageNet\_subset} & FT & 58.3 & 80.2 & 79.6 & 79.9 & 81.1 \\
 & Ours & 75.6 & 82.7 & 81.5 & 79.7 & 83.1 \\ \hline
\end{tabular}}\label{tab:diff_layers}
\end{table*}

\begin{figure}[t]
    \centering
    \includegraphics[width=0.45\textwidth]{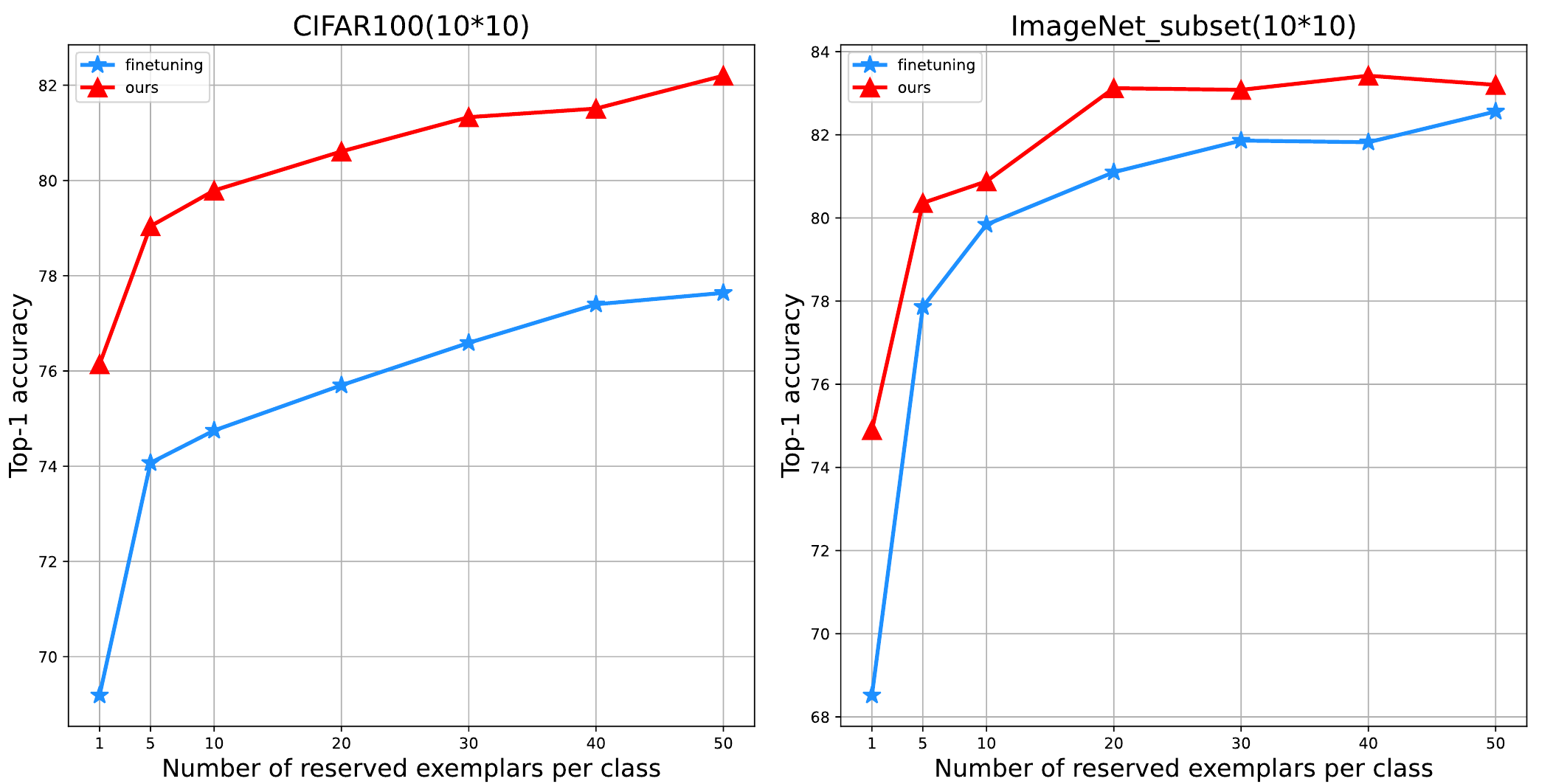}
    \caption{The comparison of accuracy when saving different number of exemplars on ImageNet\_subset.} 
    \label{fig:exem_size}
\end{figure}

\subsubsection{Validation on various vision models}

Our primary experiments employed ViT-B/16 as the image encoder. However, it is important to investigate the effectiveness of our method with other architectures as well. Hence, we validate our approach on two additional vision models: ResNet 50 and ViT-L/14. Table~\ref{tab:diff_backbones} presents the comparison between Continual-CLIP (illustrated as `C-CLIP' for short in the table) evaluation and our proposed method on ImageNet\_subset within 10 steps.  
Despite the differences in architecture and parameter count, all three image encoders exhibit significant improvements compared to Continual-CLIP evaluation. 
These results suggest that our method is effective across different scales of image encoders, irrespective of architecture.

\begin{table}[t]
\centering
\caption{Ablation study on hyper-parameters sensitivity.  (`I' stands for `ImageNet\_subset'.)}
\resizebox{\linewidth}{!}{
\begin{tabular}{c|ccccccccc}
\hline
$\lambda_{1}$    & 0.1      & 0.5       & 1        & 0.1       & 0.5       & 1       & 0.1       & 0.5       & 1              \\ \hline
$\lambda_{2}$    & 0.1      & 0.1       & 0.1      & 0.5       & 0.5       & 0.5     & 1.0       & 1.0       & 1.0            \\ \hline 
I                & 82.7     & \textbf{83.1}      & 82.5     & 80.7      & 81.2      & 81.1    & 79.7      & 79.6      & 79.5           \\ \hline
\end{tabular}}
\label{tab:hyper_sen}
\end{table}

\subsubsection{Effect of training different layers of image encoder} From Table~\ref{tab:diff_layers}, we observe that when training the last 6 blocks of the image encoder (half of the blocks in ViT-B/16), the model fails to overcome catastrophic forgetting. However, significant improvements are achieved when our proposed SG-RL and SG-KD modules are applied. Training only the linear projection layer yields decent results, but it appears that the model's potential is greatly limited with such a small number of parameters. To explore the impact of incorporating more parameters into the training process, we gradually training additional blocks. It is evident that the best performance is achieved when only the last blocks are trained.

\subsubsection{Comparison of different exemplar sizes} \label{sm:emexplar}
Some previous works~\cite{huang2016learning, masana2022class} have shown that saving a certain number of exemplars for old classes is very useful for defying forgetting. It can be seen from Fig~\ref{fig:exem_size}, the accuracy exhibits a gradual increase as the number of saved exemplars increases. 
Our proposed method improves fine-tuning results in all cases. It is worth mentioning that on ImageNet\_subset, performance of fine-tuning with only one exemplar for each old class drops a lot compared with Continual-CLIP evaluation,  
While our method still improves compared to Continual-CLIP, and shows a huge gap with fine-tuning.

\subsubsection{Hyper-parameters sensitivity}\label{section:hyper}
We conduct sensitivity analysis experiments on the two proposed losses to assess their impact. It can be seen from Table~\ref{tab:hyper_sen}, our method shows strong robustness to both hyper-parameters ranged from 0.1 to 1. The best results are obtained when $\lambda_{1}$ and $\lambda_{2}$ are set to 0.5, 0.1 respectively.

\begin{table*}[t]
\centering
\caption{Results compared to one-hot label on eight fine-grained datasets with different task splits.}
\resizebox{0.85\linewidth}{!}{
\begin{tabular}{c|c|c|c|c|c|c|c|c}
\toprule
\textbf{Dataset}     & \textbf{Food}    & \textbf{Cars}   & \textbf{Aircraft}  & \textbf{Pets}  & \textbf{Caltech}   & \textbf{Flowers}  & \textbf{CUB}   & \textbf{Dogs} \\ \midrule
split                & 11+$9 \times 10$ & $28 \times 7$   & $10 \times 10$    & 12+25  & $17 \times 6$     & $17\times6$   & $20\times10$  & $12\times10$      \\ \midrule 
One-hot Label        & 83.9    & 85.2   & 52.4     & 92.2   & 90.2     & 92.4  & 77.4     & 72.9\\
Ours                 & \textbf{90.8}    & \textbf{88.6}   & \textbf{66.6}     & \textbf{94.0}   & \textbf{92.3}     & \textbf{96.2}  & \textbf{83.7}     & \textbf{79.6} \\ \bottomrule
\end{tabular}}
\label{tab:fg_dataset2}
\end{table*}

\begin{figure*}[t]
    \centering
        \includegraphics[width=0.95\textwidth]{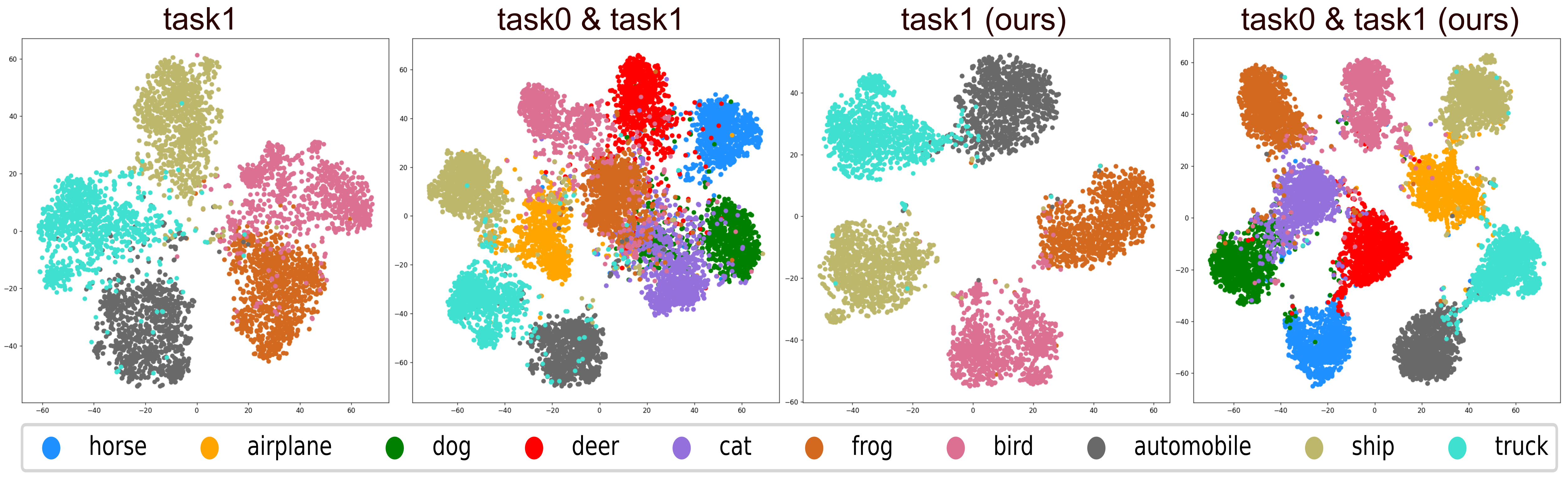}
    \caption{T-SNE visualization on test data with Continual-CLIP (left two) and our method (right two).}
    \label{fig:tsne}
\end{figure*}

\subsection{Comparison with one-hot label}

We adopted contrastive loss as the baseline performance for fine-tuning in the main paper, here we investigated the naive cross-entropy loss by one-hot label. We compute the cross-entropy loss between the predictions of the network on the current classes and the ground-truth, and use the prompt of each class as a classification head. As it can be seen from the Table~\ref{tab:fg_dataset2} and Table~\ref{tab:gen_dataset}, fine-tuning by only the cross-entropy loss calculated by one-hot label on some datasets achieve reasonable results, but it still shows a large gap compared with our method. To some extent this also reflects that only learning the image-to-label mapping tends to suffer from more severe catastrophic forgetting during incremental learning process.

\begin{table}[t]
\centering
\caption{Results compared to one-hot label on two general datasets with different task splits.}
\resizebox{0.85\linewidth}{!}{
\begin{tabular}{c|c|c}
\toprule
\textbf{Dataset}     & \textbf{ImageNet\_subset}   & \textbf{ImageNet1000} \\ \midrule
split                & $10 \times 10$              & $100 \times 10$       \\ \midrule 
One-hot Label        & 73.2                        & 64.3                   \\
Ours                 & \textbf{83.1}               & \textbf{75.1}          \\ \bottomrule
\end{tabular}}
\label{tab:gen_dataset}
\end{table}

\subsection{Visualization}
\subsubsection{Visualization of feature representation}

For the sake of simplicity and intuitiveness, we utilize the CIFAR10 dataset as an example to visualize the t-SNE representation~\cite{van2008visualizing}. We split the dataset into two steps, with each containing 5 new classes. The two sub-figures on the left shown in Fig.~\ref{fig:tsne} represent the t-SNE visualization of the feature representation evaluated on Continual-CLIP, while the right side shows the results obtained by our proposed method. For the new task (`task1'), Continual-CLIP (the first) can somewhat separate the classes, but it is evident that samples belonging to the same class are scattered and not well-clustered. Conversely, our method (the third) successfully clusters each class and significantly increases the distance between classes. When all data is mixed (`task0 \& task1'), it becomes apparent that Continual-CLIP (the second) lacks clear and meaningful boundaries between classes. Each class occupies a large space, indicating poor intra-class clustering. In contrast, ours clearly separates both the old and new tasks (the fourth). Notably, the categories \emph{`airplane'}, \emph{`truck'} and \emph{`ship'} are positioned far away from the animal categories such as \emph{`horse'}, \emph{`dog'}, \emph{`deer'}, and \emph{`cat'}. This demonstrates that our method, with its semantically-guidance, better understands the semantic meaning of the categories and capture their relationships.

\subsubsection{Heat Map Visualization}\label{sm:visualization}
\begin{figure}[t]
    \centering
    \includegraphics[width=0.48\textwidth]{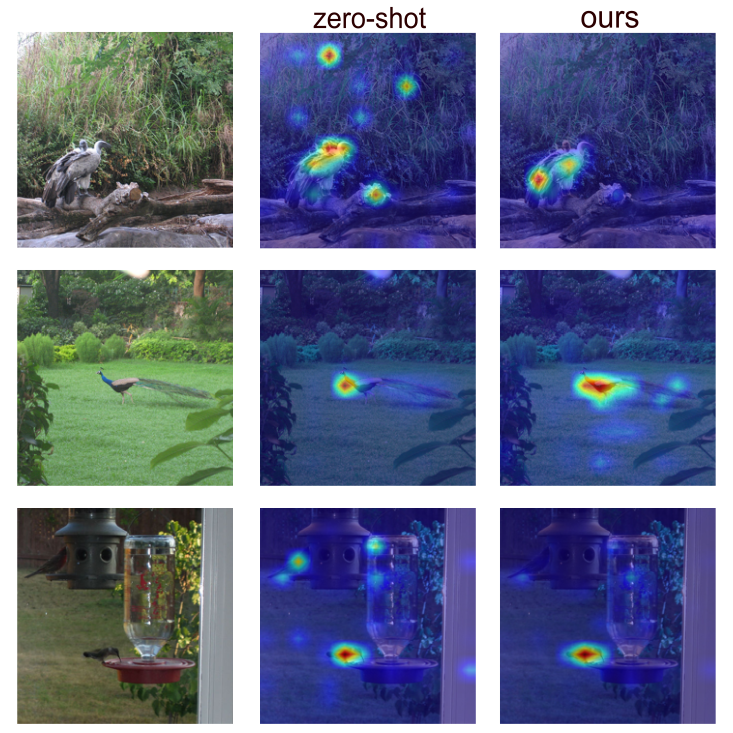}
    \caption{The heat map visualization comparison of zero-shot and our method.}
    \label{fig:attn_vis}
\end{figure}

We present the heat maps visualization in Fig.~\ref{fig:attn_vis} as it was proposed in \cite{chefer2021generic}. As can be seen from the \emph{`vulture'} in the first row, zero-shot CLIP seems to have focused part of its attention on the wood and bushes, which may indicate that it mistook these two parts for vultures, considering the color of the wood and the vultures are very similar. After applying our method, the attention is completely focused on the vulture, while the rest of the noise is completely eliminated. As for the \emph{`peacock'} in the second row, it is obvious that the attention completely covers the entire peacock after applying our method. And for the \emph{`hummingbird'} in the last row, the zero-shot CLIP seems to confuse the bird in the upper left corner, small part of the water bottle, and the hummingbird itself, while our method only focuses the attention completely on the hummingbird itself and eliminates other noise.

\section{Conclusion}
We investigated the application of large-scale vision-language pre-trained models in continual learning. 
To capture semantic similarity, we employed text embeddings from the text encoder to compute category similarity. This information was then used to generate semantically-guided label supervision, enhancing the model's understanding of category relationships during training. Additionally, we proposed a refinement technique that improved distillation loss computation by considering the semantic similarity between text embeddings of old and new classes. This approach facilitated a more precise transfer of knowledge from previous tasks to new ones. 

\minisection{Limitations.} Our method relies on the availability of text information for each task or category. In some real-world scenarios, such as image-only datasets or domains where textual descriptions are not readily available, our approach may not be directly applicable, which presents a potential avenue for future exploration.

\bibliographystyle{ieeetr}
\bibliography{main}


\begin{IEEEbiography}[{\includegraphics[width=1in,height=1.25in,clip,keepaspectratio]{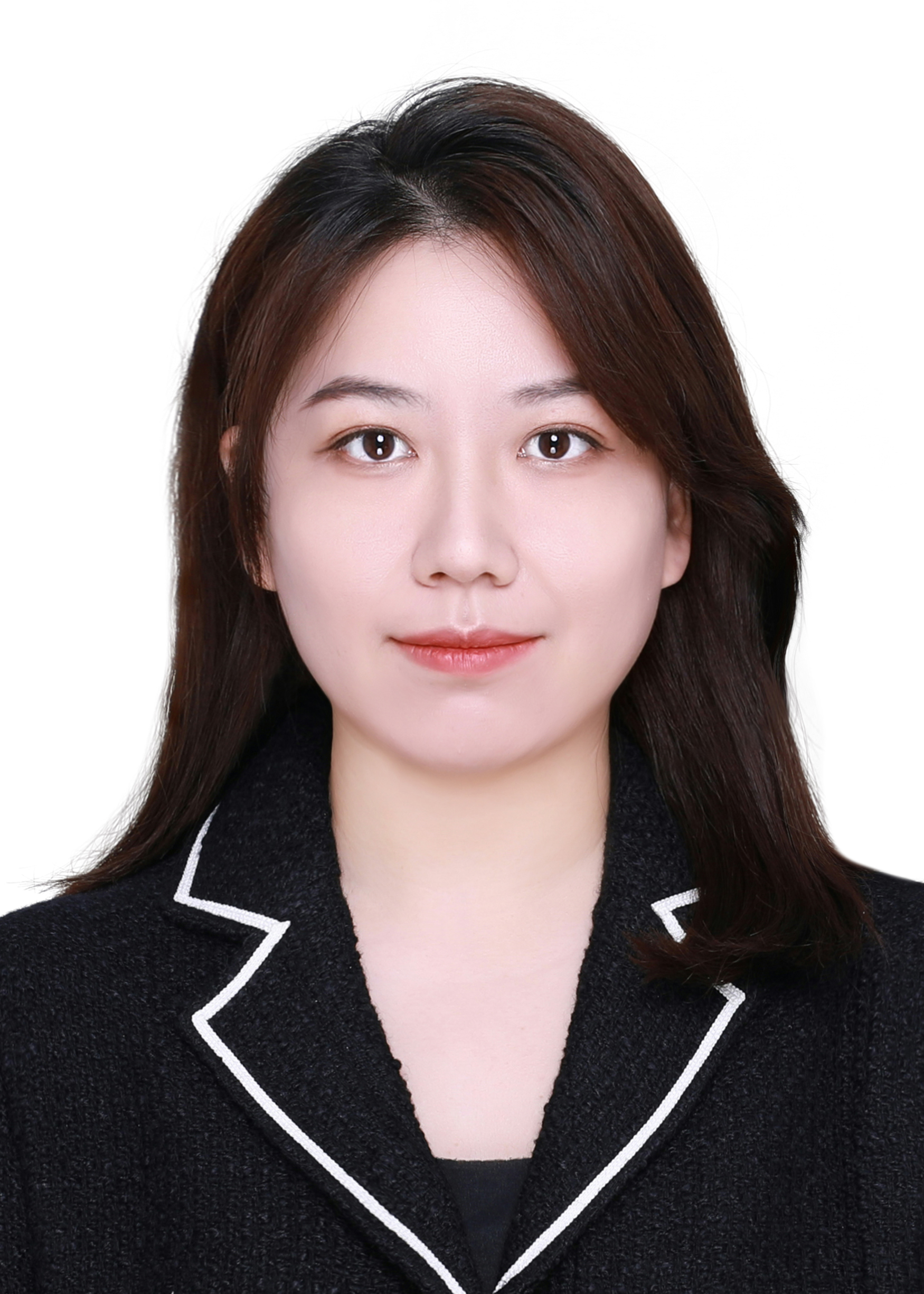}}]{Lu Yu} is currently an associate professor at Tianjin University of Technology, Tianjin, China. Before that She was a pos-doc at Heriot Watt University, Edinburgh, UK. She received her Ph.D in computer science from Autonomous University of Barcelona, Barcelona, Spain in 2019 and master degree from Northwestern Polytechnical University in 2015, Xi'an, China.  Her research interests include continual learning, metric learning, multi-model learning and color representation learning.
\end{IEEEbiography}

%\vspace{-33pt}
\begin{IEEEbiography}[{\includegraphics[width=1in,height=1.25in,clip,keepaspectratio]{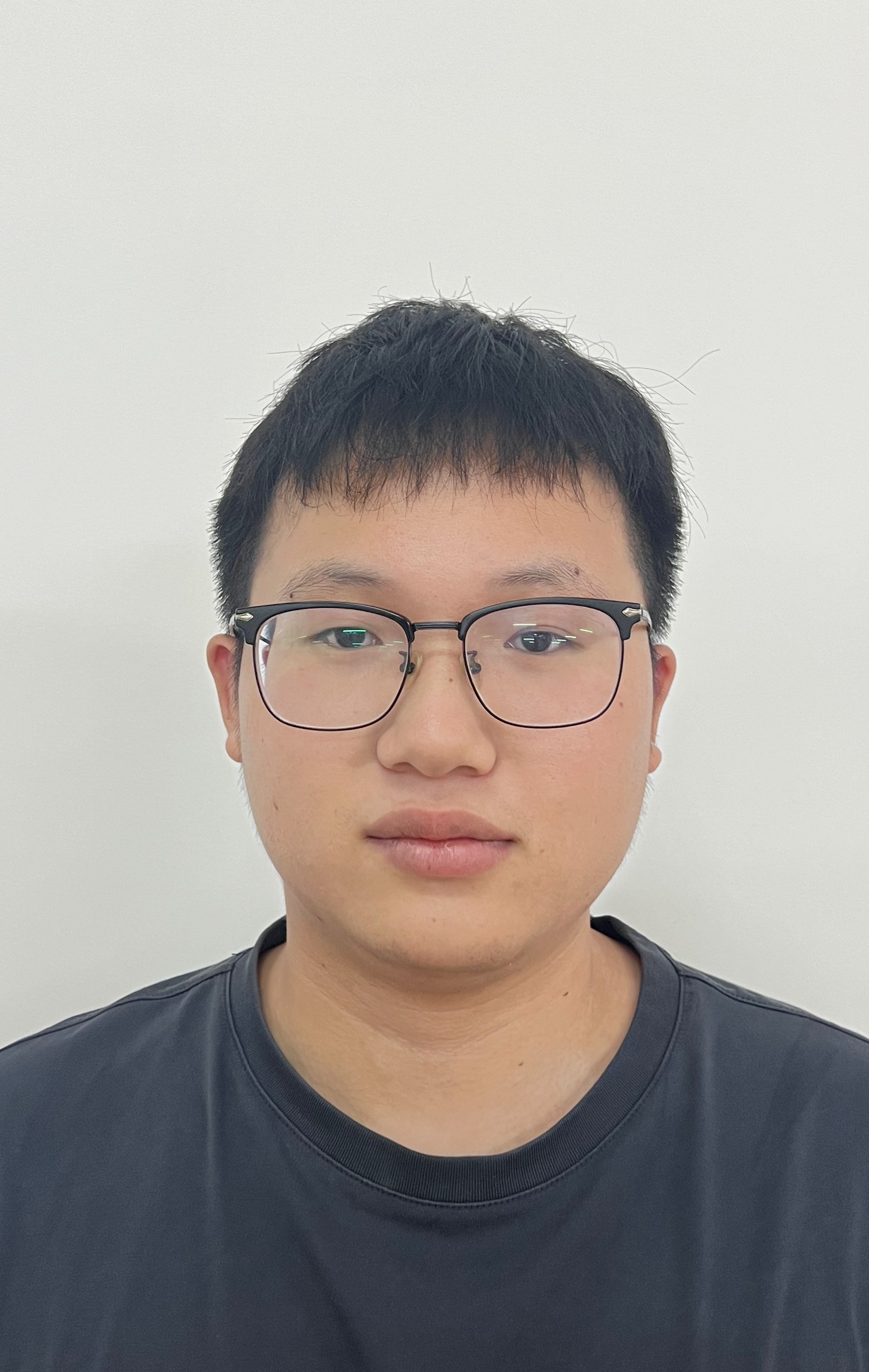}}]{Zhe Tao} received the bachelor’s degree in Urban Underground Space Engineering from Anhui Jianzhu University, Hefei, China, in 2020, and Master’s degree in Computer Technology from Jiangsu University of Science and Technology, Zhenjiang, China, in 2023. Currently, he is working toward the Ph.D. degree at Tianjin University of Technology, Tianjin, China. His research interests include deep learning and incremental learning.
\end{IEEEbiography}

\begin{IEEEbiography}[{\includegraphics[width=1in,height=1.25in,clip,keepaspectratio]{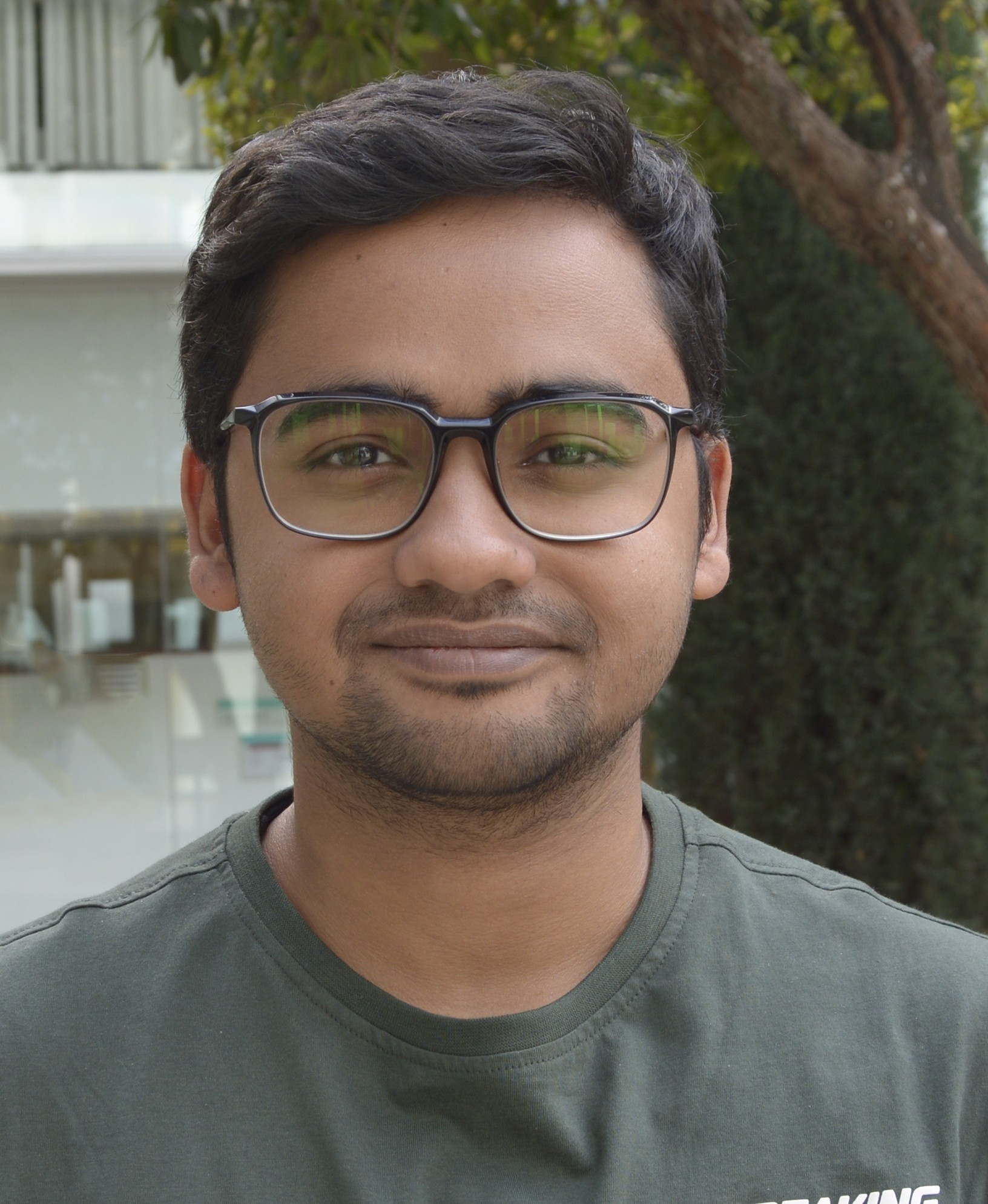}}]{Dipam Goswami} is currently a Ph.D. student at the Computer Vision Center, Universitat Autonoma de Barcelona. Before that, he graduated with a Bachelor's degree in Computer Science and a Master's degree in Mathematics from Birla Institute of Technology and Science, Pilani, India in 2022. His research interests include continual learning, federated learning, information retrieval and vision-language models.
\end{IEEEbiography}

%\vspace{-33pt}
\begin{IEEEbiography}[{\includegraphics[width=1in,height=1.25in,clip,keepaspectratio]{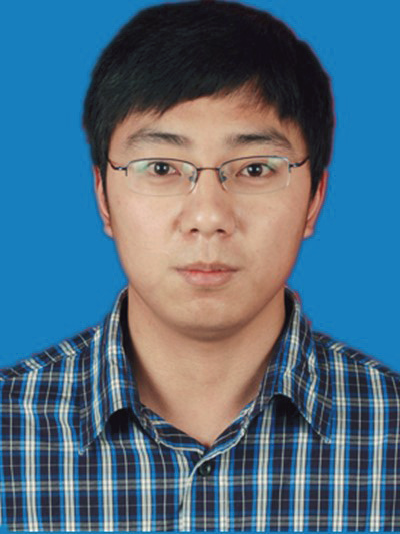}}]{Hantao Yao} (Member, IEEE) received the B.S. degree from XiDian University, Xi'an, China, in 2012. He received his Ph.D. degree in Institute of Computing Technology, University of Chinese Academy of Sciences in 2018. After graduation, he worked as a post-doctoral from 2018 to 2020 at National Laboratory of Pattern Recognition, Institute of Automation, Chinese Academy of Sciences. Now, he is an associate professor at State Key Laboratory of Multimodal Artificial Intelligence Systems, Institute of Automation, Chinese Academy of Sciences. He is the recipient of National Postdoctoral Programme for Innovative Talents. His current research interests are zero-shot learning, detection, person re-identification, and continual learning.
\end{IEEEbiography}

%\vspace{-33pt}
\begin{IEEEbiography}[{\includegraphics[width=1in,height=1.25in,clip,keepaspectratio]{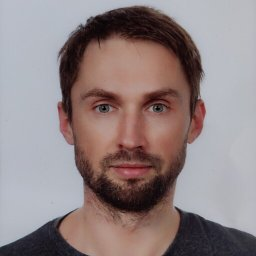}}]{Bartłomiej Twardowski} is a Research Group Leader at the IDEAS Research Institute in Poland and a researcher at the Computer Vision Center, Universitat Autònoma de Barcelona. He earned his Ph.D. in 2018, specializing in recommender systems and neural networks. After completing his doctorate, he worked as an assistant professor in the AI group at Warsaw University of Technology for 1.5 years before joining the Computer Vision Center at UAB for a postdoctoral fellowship. He has been actively involved in numerous research projects in deep learning, natural language processing, and machine learning, with funding ranging from €40k to €1.4M, and is a recipient of the prestigious Ramón y Cajal fellowship. 
\end{IEEEbiography}

%\vspace{-33pt}
\begin{IEEEbiography}[{\includegraphics[width=1in,height=1.25in,clip,keepaspectratio]{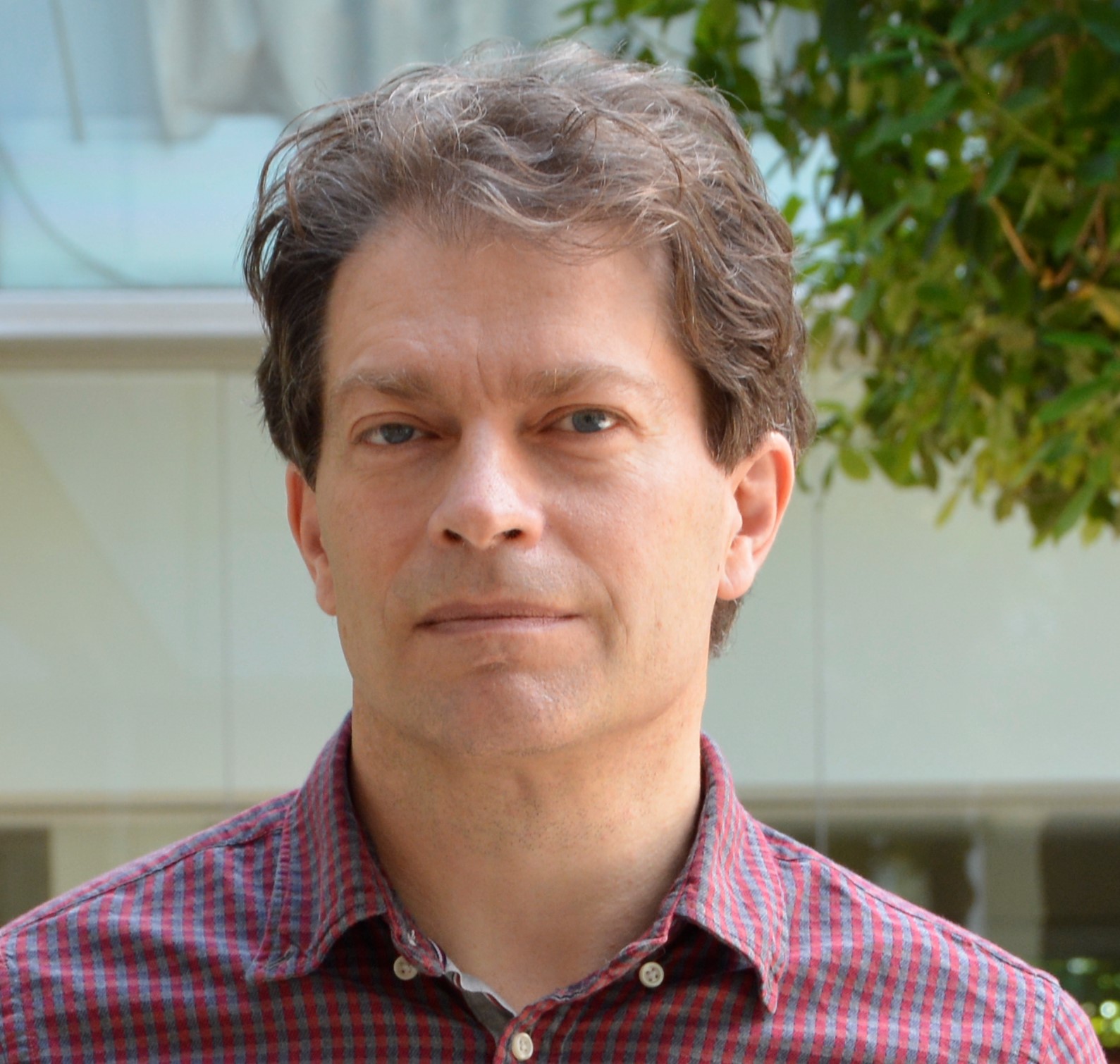}}]{Joost van de Weijer} received the Ph.D. degree from the University of Amsterdam in 2005. He was a Marie Curie Intra-European Fellow at INRIA Rhone-Alpes, France, and a Ramon y Cajal Fellow at the Universitat Autònoma de Barcelona, Spain, where he is currently the leader of the Learning and Machine Perception (LAMP) Team at the Computer vision Center. His main research directions are continual learning, domain adaptation and generative vision models. He is ELLIS fellow.
\end{IEEEbiography}

%\vspace{-33pt}
\begin{IEEEbiography}[{\includegraphics[width=1in,height=1.25in,clip,keepaspectratio]{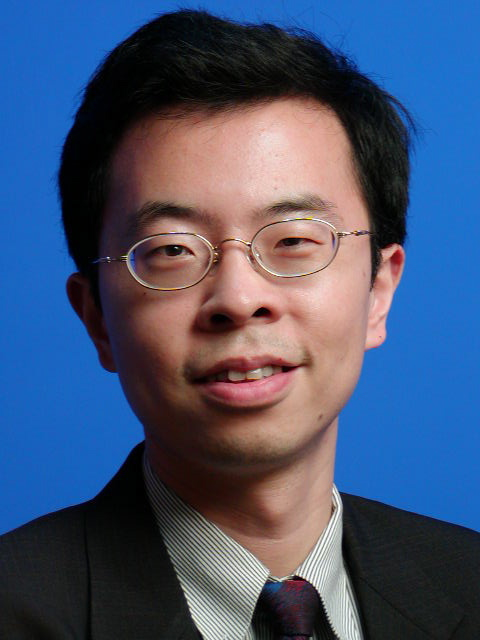}}]{Changsheng Xu} (M’97–SM’99–F’14) is a Professor in State Key Laboratory of Multimodal Artificial Intelligence Systems (MAIS), Institute of Automation, Chinese Academy of Sciences. His research interests include multimedia content analysis/indexing/retrieval, pattern recognition and computer vision. He has hold 50 granted/pending patents and published over 400 refereed research papers in these areas. Dr. Xu has served as associate editor, guest editor, general chair, program chair, area/track chair and TPC member for over 20 IEEE and ACM prestigious multimedia journals, conferences and workshops, including IEEE Trans. on Multimedia, ACM Trans. on Multimedia Computing, Communications and Applications and ACM Multimedia conference. He is IEEE Fellow, IAPR Fellow and ACM Distinguished Scientist.
\end{IEEEbiography}

\appendices
\clearpage

\section{Discussions with RanPAC}\label{comb}
When combined with a RanPAC-style classifier, our model consistently outperforms the original RanPAC method, demonstrating the robustness and transferability of the learned features. Under an oracle setting—where RanPAC’s classifier is constructed using full data statistics—our approach achieves even higher performance, further indicating that the feature representations learned by our method possess strong linear separability.

\begin{table*}[]
\centering
\caption{Combination with RanPAC-style classifier.  
$\dagger$ denotes RanPAC classifier adaptation using exemplars with continually updated feature extractor.}
\resizebox{0.8\textwidth}{!}{
\begin{tabular}{c|c|c|c|c|c}
\toprule
\textbf{Dataset}                                               & \textbf{CIFAR100}   & \textbf{Aircraft} & \textbf{Cars}     & \textbf{Dogs}   & \textbf{ImageNet-R} \\ \midrule
\#classes                                                      & 100                 & 100               & 196               & 120             & 200                 \\ 
split                                                          & $10 \times 10$      & $10 \times 10$    & $28 \times 7$     & $12 \times 10$  & $20 \times 10$      \\ \midrule
RanPAC \cite{mcdonnell2024ranpac}                              & 82.0                & 62.3              & 86.3              & 77.0            & 78.0       \\  

Ours $+$ RanPAC$^\dagger$ \cite{mcdonnell2024ranpac} & 84.8                & 69.3              & 88.9              & 78.5            & 82.4       \\ 
Ours $+$ Oracle RanPAC \cite{mcdonnell2024ranpac}         & 86.6                & 71.4              & 91.9              & 81.5            & 83.8       \\ 
\bottomrule
\end{tabular}}
\label{tab:com_ranpac}
\end{table*}

\section{Evaluation on Multi-Domain Task Incremental Learning Setting}\label{MTIL}
The Multi-Domain Task Incremental Learning (MTIL) setting was first introduced by ZSCL\cite{zheng2023preventing}, which proposed a knowledge distillation module and a weight ensemble mechanism to mitigate zero-shot performance degradation. MTIL focuses on continual learning across different datasets, with an emphasis on preserving zero-shot capabilities. In contrast, our class-incremental learning setting aligns with prior works, aiming to continually learn new tasks while retaining knowledge of previously learned classes. 

We compared our method with these methods mentioned above in Table~\ref{tab:mtil_order_I} and Table~\ref{tab:mtil_order_II} following the same datasets order I and order II in ZSCL. The backbones of all methods are CLIP ViT-B/16 for fair comparison.  
Under MTIL setting, methods are evaluated with \textit{Transfer}, \textit{Last} and \textit{Average} three main metrics, corresponding to zero-shot performance, averaged accuracy after last task and averaged incremental accuracy respectively. As shown in Table~\ref{tab:mtil_order_I}, our method achieves optimal or second-best average performance across all evaluation metrics. Notably, it yields a 1.2\% improvement on the \textit{Average} metric, while being only 0.3\% and 0.2\% lower than TreeProbe on the \textit{Last} and \textit{Transfer} metrics, respectively. The results for Order II are presented in Table~\ref{tab:mtil_order_II}, where our method consistently achieves superior average performance across all evaluation metrics. 

\begin{table*}[]
    \centering
    \caption{Comparison with different methods on the MTIL benchmark (Order I) in terms of `Transfer', `Last', and `Average' performance(\%).}
    \resizebox{\textwidth}{!}{
    \begin{tabular}{c c c c c c c c c c c c c c}
        \toprule
        &\multirow{-4}{*}{\textbf{\large{Methods}}} & \rotatebox{90}{Aircraft} & \rotatebox{90}{Caltech101} & \rotatebox{90}{CIFAR100} & \rotatebox{90}{DTD} & \rotatebox{90}{EuroSAT} & \rotatebox{90}{Flowers} & \rotatebox{90}{Food} & \rotatebox{90}{MNIST} & \rotatebox{90}{OxfordPets} & \rotatebox{90}{Cars} & \rotatebox{90}{SUN397} & \multirow{-4}{*}{\textbf{\large{Average}}} \\
        \midrule
        \multirow{4}{*}{\textit{Transfer}} 
        & ZSCL \cite{zheng2023preventing} &- & 86.0 & 67.4 & 45.4 & 50.4 & 69.1 & 87.6 & \textbf{61.8} & 86.8 & 60.1 & \textbf{66.8} & 68.1\\
        & MoE4CLIP \cite{yu2024boosting} &- & \textbf{87.9} & \textbf{68.2} & 44.4 & 49.9 & 70.7 & 88.7 & 59.7 & 89.1 & 64.5 & 65.5 & 68.9\\
        & TreeProbe \cite{zhucontinual} &- & \textbf{87.9} & \textbf{68.2} & 45.3 &  54.6 & 71.1 & \textbf{88.9} & 59.5 & 89.1 & \textbf{64.6} & 64.1 & \textbf{69.3}\\
        & Ours & -& 84.3 & 68.0 & \textbf{46.1} & \textbf{56.0} & \textbf{71.3} & 87.0  & 57.9 & \textbf{91.0} & \textbf{64.6}  & 65.0  & \underline{69.1}  \\
        
        \midrule
        \multirow{4}{*}{\textit{Last}} % \textit{Last} & & & & & & & & & & &\\
        & ZSCL \cite{zheng2023preventing} & 40.6 & 92.2 & 81.3 & 70.5 & 94.8 & 90.5 & 91.9 & \textbf{98.7} & \textbf{93.9} & 85.3 & 80.2 & 83.6\\
        & MoE4CLIP \cite{yu2024boosting} & 49.8 & 92.2 & \textbf{86.1} &\textbf{ 78.1} & \textbf{95.7} & 94.3 & 89.5 & 98.1 & 89.9 & 81.6 & 80.0 & 85.0\\
        & TreeProbe \cite{zhucontinual} & 52.5 & \textbf{95.6} & 81.9 & 66.9 & 95.6 & \textbf{95.6} & \textbf{92.2} & 98.6 & 93.2 & \textbf{86.2} & \textbf{81.8} & \textbf{85.5}\\
        & Ours & \textbf{64.5} & 93.6 & 79.8 & 70.8 & 93.2  & 94.9  & 91.1  & 98.0  & 92.9  & 81.1  & 77.6 & \underline{85.2}\\
        
        \midrule
        \multirow{4}{*}{\textit{Average}}
        & ZSCL \cite{zheng2023preventing} & 45.1 & 92.0 & 80.1 & 64.3 & 79.5 & 81.6 & 89.6 & \textbf{75.2} & 88.9 & 64.7 & \textbf{68.0} & 75.4\\
        & MoE4CLIP \cite{yu2024boosting} & 50.2 & 91.9 & \textbf{83.1} & \textbf{69.4} & 78.9 & 84.0 & 89.1 & 73.7 & 89.3 & 67.7 & 66.9 & \underline{76.7}\\
        & TreeProbe \cite{zhucontinual} & 52.5 & \textbf{93.6} & 79.5 & 61.9 & \textbf{80.5} & 78.0 & \textbf{90.4} & 73.7 & 90.2 & \textbf{68.5} & 65.7 & 75.9 \\
        & Ours & \textbf{66.6} & 93.3 & 79.3 & 65.8 & 80.3 & \textbf{84.2} & 89.2 & 72.6 & \textbf{91.8} & 67.9 & 66.2 & \textbf{77.9}\\
        \bottomrule
    \end{tabular}
    }
    \label{tab:mtil_order_I}
\end{table*}

\section{Results with Learnable Backbone}\label{LB}
In our main experimental table\ref{tab:cifar_3}, methods are distinguished by whether they employ a learnable backbone (LB), indicating if pre-trained weights are updated during continual learning. A significant portion of existing state-of-the-art methods operate with a frozen backbone. Our proposed approach, in contrast, involves fine-tuning part of the pre-trained weights of the Vision Transformer. This naturally raises a question regarding the source of our observed performance gains: to what extent are they attributable to our novel semantic guidance techniques (i.e., SG-KD and SG-RL) versus the general benefits derived from fine-tuning a segment of the backbone?
To address this and provide a more direct comparison of the core methodological contributions, we have included an additional comparative experiment in Table\ref{tab:lb_dataset}. In this supplementary analysis, we adapted several leading competitor methods, which originally used a frozen backbone, to also fine-tune the last residual block of their ViT backbone, mirroring the parameter update strategy of our method. While these adapted competitor methods predictably showed performance improvements over their frozen-backbone counterparts due to the increased learnability, our proposed method consistently maintained a significant advantage even under this condition. This finding strongly suggests that the primary driver of our method's superior performance stems from the effective exploitation of semantic knowledge via our proposed SG-KD and SG-RL mechanisms, rather than being solely an artifact of partial backbone fine-tuning. This isolates and underscores the distinct benefits contributed by our novel techniques.

\section{More Details} \label{re_details}
\subsection{Unified Exemplar Management for Replay}
Among these compared methods, some of them (L2P, DualPrompt, RanPAC, CODA-Prompt, ConvPrompt, EASE, and RAPF) were originally proposed as exemplar-free CIL algorithms, some were exemplar-based CIL methods (SPU, CLAP, PROOF). CoOp and CoCoOp were proposed for prompt learning in a non-continual manner. Except EASE and RAPF, for all of these compared methods, we employed the herding algorithm to select and store 20 exemplars per old class.\footnote{EASE employs prototype synthesis strategy to construct classifier, RAPF employs a decomposed parameter fusion strategy for its adapter module, these two methods are not well compatible with exemplars replay.} These exemplars were then replayed during the training of subsequent incremental tasks, and their loss was incorporated into the total training objective for each method. For CoOp and CoCoOp, we adapted them to the CIL setting and also applied the same exemplar replay strategy.

\subsection{Hyperparameter and Training Parameter Settings}
For each baseline method, we prioritized using the optimal hyperparameters and training configurations reported in their respective original publications whenever possible.
If the original parameters were not directly applicable to our specific CIL setup or datasets, or if complete details were not provided, we conducted a reasonable hyperparameter search (e.g., on a validation set) to identify settings that yield the best performance for each method under our evaluation framework.

By adhering to these careful implementation practices and standardized evaluation protocols, we aimed to establish a robust and equitable comparison platform for all methods, ensuring that the observed performance differences accurately reflect their capabilities in the CIL context.

\begin{table*}[]
    \centering
    \caption{Comparison with different methods on the MTIL benchmark (Order II) in terms of `Transfer', `Last', and `Average' performance(\%). Results of TreeProbe on order II are reproduced with their official published code, as this experiment was not reported in the original paper.}
    \resizebox{\textwidth}{!}{
    \begin{tabular}{c cc c c c c c c c c c c c}
        \toprule
        &\multirow{-4}{*}{\textbf{\large{Methods}}} & \rotatebox{90}{Cars} & \rotatebox{90}{Food} & \rotatebox{90}{MNIST} & \rotatebox{90}{OxfordPets} & \rotatebox{90}{Flowers} & \rotatebox{90}{SUN397} & \rotatebox{90}{Aircraft} & \rotatebox{90}{Caltech101} & \rotatebox{90}{DTD} & \rotatebox{90}{EuroSAT} & \rotatebox{90}{CIFAR100} & \multirow{-4}{*}{\textbf{\large{Average}}} \\
        \midrule
        \multirow{4}{*}{\textit{Transfer}} 
        & ZSCL \cite{zheng2023preventing} &- & 88.3 & 57.5 & 84.7 & 68.1 & 64.8 & 21.1 & 88.2 & 45.3 & \textbf{55.2} & 68.2 & 64.1 \\
        & MoE4CLIP \cite{yu2024boosting} &- & \textbf{88.8} & \textbf{59.5} & 89.1 & 69.9 & 64.6 & 18.1 & 86.9 & 43.7 & 54.6 & 68.2 & \underline{64.3}\\
        & TreeProbe* \cite{zhucontinual} & - & 87.9 & \textbf{59.5} & 89.1 & 71.1 & 61.1 & \textbf{24.7} & 87.2 & 43.5 & 54.7 & 62.7 & 64.2\\
        & Ours &- & 87.0 & 58.1 & \textbf{90.8} & \textbf{71.4} & \textbf{65.1} & 23.8 & \textbf{88.5} & \textbf{46.4} & 55.1 & \textbf{68.3} & \textbf{65.5} \\
        
        \midrule
        \multirow{4}{*}{\textit{Last}} 
        & ZSCL \cite{zheng2023preventing} & 78.2 & 91.1 & 97.6 & 92.5 & 87.4 & 78.2 & 45.0 & 92.3 & 72.7 & \textbf{96.2} & 86.3 & 83.4\\
        & MoE4CLIP \cite{yu2024boosting} & 84.1 & 88.5 & 94.0 & 91.8 & 94.1 & 77.8 & 50.4 & \textbf{93.3} & \textbf{77.1} & 87.7 & \textbf{86.6} & \underline{84.2}\\
        & TreeProbe* \cite{zhucontinual} & 86.3 & \textbf{92.2} & \textbf{98.5} & 91.5 & 81.0 & \textbf{79.0} & 49.7 & 90.2 & 62.7 & 88.2 & 81.9 & 81.9\\
        & Ours & \textbf{86.6} & 90.6 & 96.2 & \textbf{93.1} & \textbf{94.9} & 74.0 & \textbf{57.2} & 93.2 & 69.4 & 94.7 & 82.2 & \textbf{84.7}\\
        
        \midrule
        \multirow{4}{*}{\textit{Average}} 
        & ZSCL \cite{zheng2023preventing} & 81.7 & 91.3 & 91.1 & 91.0 & 82.9 & \textbf{72.5} & 33.6 & 89.7 & 53.3 & \textbf{62.8 }& \textbf{69.9} & 74.5 \\
        & MoE4CLIP \cite{yu2024boosting} & 84.9 & 89.9 & 89.3 & 91.4 & 86.2 & 72.2 & 33.4 & 89.4 & 53.3 & 61.4 & \textbf{69.9} & \underline{74.7}\\
        & TreeProbe* \cite{zhucontinual} & 86.3 & \textbf{91.8} & \textbf{91.4} & 91.1 & 83.1 & 71.7 & 36.0 & 89.2 & 49.8 & 60.9 & 64.5 & 74.3\\
        & Ours & \textbf{88.2} & 91.0 & 90.4 & \textbf{93.0} & \textbf{86.4} & 71.1 & \textbf{39.7} & \textbf{90.5} & \textbf{53.6} & 62.6 & 69.6 & \textbf{76.0}\\
        \bottomrule
    \end{tabular}
    }
    \label{tab:mtil_order_II}
\end{table*}

\begin{table*}[t]
\centering
\caption{
Results with a learnable backbone across multiple evaluated datasets.
}
\resizebox{1\linewidth}{!}{
\begin{tabular}{c|c|c|c|c|c|c|c|c|c|c|c|c}
\toprule
\textbf{Dataset}        & \textbf{LB}  & \textbf{CIFAR100} & \textbf{ImageNet\_sub} & \textbf{ImageNet1000} & \textbf{Food}    & \textbf{Cars}   & \textbf{Aircraft}  & \textbf{Pets}  & \textbf{Caltech}   & \textbf{Flowers}  & \textbf{CUB}   & \textbf{Dogs} \\ \midrule
\#classes                &              & 100               & 100                      & 1000                  & 101     & 196    & 100      & 37     & 102      & 102     & 200    & 120                                                                           \\ 
split                    &            & $10 \times 10$    & $10 \times 10$           & $100 \times 10$       & 11+$9 \times 10$ & $28 \times 7$   & $10 \times 10$    & $12+25$  & $17 \times 6$     & $17\times6$   & $20\times10$  & $12\times10$                \\ \midrule 
CoOp\cite{zhou2022learning}          & \checkmark   & 68.1   & 70.2     & 66.7    & 84.6    & 86.1   & 57.6     & 94.0   & 90.2     & 97.0  & 80.1  & 75.8  \\
CoCoOp\cite{zhou2022conditional}     & \checkmark   & 65.9   & 71.0     & 56.0    & 79.1    & 84.9   & 59.8     & 92.8   & 88.2     & \textbf{97.2}  & 77.0  & 69.7 \\
L2P\cite{wang2022learning}           & \checkmark   & 65.9   & 72.7     & 63.3    & 81.7    & 86.9   & 61.1     & 92.8   & 88.6     & 97.0  & 82.6  & 75.9  \\
DualPrompt\cite{wang2022dualprompt}  & \checkmark   & 65.1   & 68.4     & 64.0    & 80.5    & 82.7   & 45.5     & 90.6   & 86.4     & 96.7  & 79.6  & 70.3   \\
Ours                                 & \checkmark   & \textbf{83.3}     & \textbf{83.1}     & \textbf{75.1}         & \textbf{90.9}    & \textbf{88.6}   & \textbf{66.6}     & \textbf{94.0}   & \textbf{92.3}     & 96.2  & \textbf{83.7}     & \textbf{79.6}    \\ \bottomrule
\end{tabular}}
\label{tab:lb_dataset}
\end{table*}

\begin{itemize}
    \item \textbf{CoOp} We adopted the class-specific prompts variant due to its superior performance. The number of context words (i.e., the number of learnable prompt tokens) was set to 16 per class. The learning rate was set to 0.01, batch size was 256, and each task was trained for 10 epochs.
    \item \textbf{CoCoOp} We set the number of prompt tokens to 16. The meta-network consisted of two linear layers with a ReLU activation function between them. The learning rate was 0.001, the batch size was 8, and training spanned 10 epochs per task.
    \item \textbf{L2P} We set the prompt length to 5 and prompt pool size equal to the number of tasks, meaning each task had a corresponding set of prompts. The trade-off for loss between queries and learnable keys was 0.1. A learning rate of 0.03 and a batch size of 16 were used. Standard datasets were trained for 5 epochs per task, extended to 10 epochs for fine-grained datasets to ensure convergence.
    \item \textbf{DualPrompt} We set the length of general prompts to 5 and applied them to the first two transformer layers, while expert prompts, also of length 5, were used in the subsequent three layers. The prompt pool size was set equal to the number of tasks. The learning rate was 0.03 with a batch size of 24. General datasets were trained for 5 epochs per task, extended to 10 epochs for fine-g                            rained datasets.
    \item \textbf{RanPAC} We set $M$ to 10,000, which is utilized to construct the ramdom matrix for mapping features to high-dimensional space.
    \item \textbf{CODA-Prompt} We set the length of prompt to 8, the prompt pool size to match the number of classe, the feature dimension of the prompts is 768. We set the learning rate to 0.001 and the batch size to 128, training for 20 epochs per task. We used the Adam optimizer with a cosine learning rate scheduler.
    \item \textbf{ConvPrompt} We configured a kernel size of 17. It learned 5 prompts per task, each of length 20, and the prompt pool size was equal to the number of tasks. Only expert prompts were utilized, applied in the first 7 transformer layers.
    \item \textbf{EASE} For CIFAR100, the learning rate was 0.025, batch size 48, and weight decay 0.0005. Other datasets adopted the ImageNet-R configuration: learning rate 0.05, batch size 16, and weight decay 0.005. All datasets were trained for 20 epochs per task using the SGD optimizer with a cosine annealing scheduler. The adapter's intermediate dimension was 64, and the ensembling coefficient $\alpha$ between different subspaces was 0.1.
    \item \textbf{SPU} We set the selection rate to 0.1 and the percentage of samples per task used for calculating parameter importance to 0.25. The learning rate was 7.5e-6, the batch size was 64, and the weight decay was 0.2. We employed the AdamW optimizer with a cosine annealing scheduler.
    \item \textbf{CLAP} the first stage training used a learning rate of 0.001, a batch size of 32, and the SGD optimizer with a cosine annealing scheduler for 5 epochs per task. The second stage (class-balanced consolidation) trained for 2 epochs with a learning rate of 0.0001. The number of Monte Carlo forward passes sampled from probabilistic adapters was 20. A single-layer transformer-decoder was used in the VGA Module, and task-specific probabilistic adapters were learned for each task. The coefficient $\lambda$ for the prior-matching DKL term was 0.001, and $\gamma$ for the past-task distribution regularization loss was 15.
    \item \textbf{PROOF} We use single linear layer for the task-specific MLP for each task. The initial learning rate was set to 0.001, the batch size to 64, and the weight decay to 0.05. We utilized the SGD optimizer with a cosine annealing scheduler.
\end{itemize}

For methods based solely on a vision encoder (L2P, DualPrompt, RanPAC, CODA-Prompt, ConvPrompt, and EASE), we utilized the vision features directly for classification. This involved discarding the projection layer from the pre-trained CLIP weights that is originally used to map vision features to the language space. Conversely, for methods designed for or leveraging vision-language models (SPU, CLAP, RAPF, and PROOF, as well as our adaptations of CoOp and CoCoOp), we employed all pre-trained weights of the CLIP model.

% {\color{blue}{
% % use section* for acknowledgment
% \ifCLASSOPTIONcompsoc
%   % The Computer Society usually uses the plural form
%   \section*{Acknowledgments}
% \else
%   % regular IEEE prefers the singular form
% \fi


\ifCLASSOPTIONcaptionsoff
  \newpage
\fi

\end{document}